%% file: acl_latex.tex
\documentclass[11pt]{article}

\usepackage[preprint]{acl}

\usepackage{times}
\usepackage{latexsym}

\usepackage[T1]{fontenc}

\usepackage[utf8]{inputenc}

\usepackage{microtype}

\usepackage{inconsolata}

\usepackage{graphicx}
\usepackage{import}   
\usepackage{booktabs} 
\usepackage{array}    
\usepackage{amsmath}  
\usepackage{amssymb}  
\usepackage{placeins} 
\usepackage{fontawesome5}
\usepackage{xurl}

\newcommand{\llama}{\texttt{LLaMA~3.1~8B}}
\newcommand{\qwen}{\texttt{Qwen~2.5~7B}}
\newcommand{\modelwithlogo}[2]{%
  \IfFileExists{#1}{\raisebox{-0.15em}{\includegraphics[height=0.95em]{#1}}\hspace{0.3em}}{}#2}

\definecolor{tabGroup}{HTML}{6D6C6C}  
\definecolor{tabRule}{HTML}{C4C4C2}   
\definecolor{deltaUp}{HTML}{2B9C8D}   
\definecolor{deltaDown}{HTML}{D1495B} 
\newcommand{\tabsans}{\fontfamily{phv}\fontsize{8}{9.5}\selectfont}
\newcommand{\colhead}[1]{{\tabsans #1}}
\newcommand{\taskgroupn}[2]{%
  \multicolumn{#1}{@{}l@{}}{\rlap{\makebox[\linewidth][l]{%
    \tabsans\color{tabGroup}#2\hspace{0.5em}%
    {\color{tabRule}\leaders\hrule height \dimexpr0.6ex+0.2pt\relax depth \dimexpr-0.6ex+0.2pt\relax\hfill}}}}}
\newcommand{\taskgroup}[1]{\taskgroupn{9}{#1}}
\newcommand{\groupskip}{\addlinespace[0.65em]}
\newcommand{\dpos}[1]{\textcolor{deltaUp}{#1}}
\newcommand{\dneg}[1]{\textcolor{deltaDown}{\textminus#1}}
\newlength\tabdeltawd
\newlength\tabmodelgap

\title{Chain-of-Thought Hidden States Reveal\\Task-Solving Directions for Direct Answers}

\author{Houman Mehrafarin$^{1,2}$ \quad Ioannis Konstas$^{1}$ \\
  $^{1}$Heriot-Watt University, Edinburgh, United Kingdom \\
  $^{2}$University of Edinburgh, Edinburgh, United Kingdom \\
  \texttt{\{h.mehrafarin, i.konstas\}@hw.ac.uk}}

\begin{document}
\maketitle

\begin{abstract}
Chain-of-Thought (CoT) prompting improves the performance of Large Language Models by eliciting intermediate reasoning steps. However, CoT is far from perfect: on some tasks, models perform worse than with direct-answer prompting, and on others they generate a few hundred tokens to reach the correct answer. We start by labelling each CoT hidden state as \emph{recovering} if patching it into the direct prompt yields the correct answer, and \emph{failed} otherwise. A linear probe shows that the model encodes these two classes differently. We then show that the representation space of the direct prompt holds a direction for each question that leads the model, in a short generation, to the correct answer more often than CoT does. Specifically, we rotate the direct prompt's last hidden state once, before generation, towards a direction built for each question from its own recovering and failed states. Across 25 tasks, these directions raise the direct prompt's mean accuracy from 46\% to 85\% on \llama{} (CoT:~69\%) and from 52\% to 86\% on \qwen{} (CoT:~77\%), with a tenth of CoT's tokens. CoT is thus not the shortest route to this accuracy, and these directions can be used in future work to train models to align their states with them.\footnote{Code: \url{https://github.com/hmehrafarin/hidden-state-directions}}
\end{abstract}

\begin{figure*}[t]
  \centering
  \includegraphics[width=\textwidth]{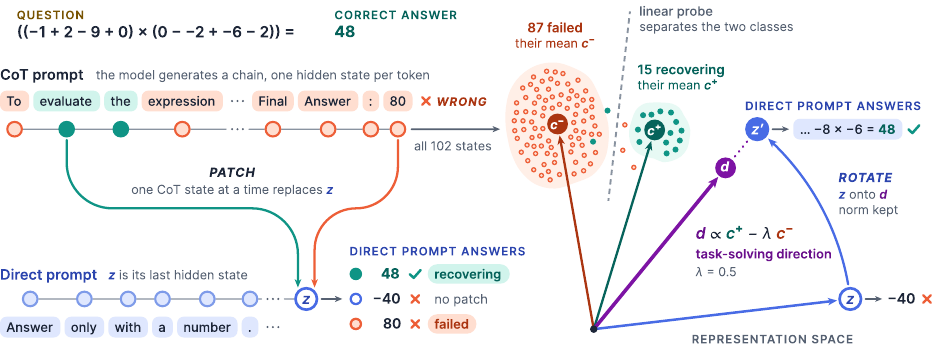}
    \caption{Overview of our analysis on one Multistep Arithmetic question. Each CoT state is patched in place of the last state $z$ of the direct prompt and labelled \emph{recovering} if the model then answers correctly and \emph{failed} otherwise. The failed states and recovering states occupy different parts of the representation space, so that a linear probe predicts this label from a single state. The means of the question's recovering and failed states, $c^{+}$ and $c^{-}$, give the task-solving direction $d \propto c^{+} - 0.5\,c^{-}$. Before generation, we rotate $z$ onto the direction of $d$, keeping its norm; the direct prompt then writes four short lines and answers $48$ instead of $-40$, even though the CoT itself answers $80$.}
  \label{fig:overview}
\end{figure*}

\section{Introduction}

Chain-of-Thought (CoT) prompting lets Large Language Models (LLMs) answer questions that require several steps of inference by generating those steps before the answer \citep{wei2022chain,kojima2022large}. The chain often runs to a few hundred tokens per question and can still end in a wrong answer. Its benefit falls mainly on mathematical and symbolic tasks \citep{sprague2025cot}, and on some tasks it lowers accuracy below that of answering directly \citep{liu2025mind}.

\citet{whencotfails} provide causal evidence that when the chain ends in a wrong answer, the information needed to solve the task is often still present in its hidden states. Using activation patching on GSM8K \citep{gsm8k}, they show that some hidden states of the chain, patched into a run of the same question prompted to answer directly, make that run produce the correct answer, which reveals a discrepancy between what the model generates and what it computes internally. We call the states whose patch makes the direct run answer correctly \emph{recovering}, and the rest \emph{failed}. Their analysis, however, covers only arithmetic reasoning and stops at detecting this information. It does not examine whether recovering states share characteristics that systematically distinguish them from failed ones, nor does it use the patch outcomes to reduce this discrepancy.

We instead use these outcomes as labels, first to test whether the two can be told apart and then to reduce this discrepancy. We do so for \llama{} and \qwen{} on GSM8K, StrategyQA \citep{strategyqa} and 23 subtasks of BIG-Bench Hard \citep{bbh}. These 25 tasks span six reasoning types: arithmetic, logical, temporal, spatial, causal and linguistic. For each task, we train a linear probe to predict whether a CoT hidden state is recovering or failed. The probe separates the two classes on held-out questions, so the label is linearly decodable from the hidden state alone. A probe trained on one task also separates them on other tasks, most of all on tasks of the same reasoning type. The probes show that the model encodes recovering and failed states differently. The two classes thus have distinct means in the representation space, which gives us the grounding to build a direction from these means.

To reduce the discrepancy, we build a contrastive direction for each question from the mean of the recovering states and the mean of the failed states of its own CoT chain, and refer to it as the question's \emph{task-solving direction}. This direction is the recovering mean minus a fraction of the failed mean, which keeps it close to the recovering mean and far from the failed mean (Figure~\ref{fig:overview}). It differs from the standard difference of means \citep{marks2024geometry,rimsky-etal-2024-steering,arditi2024refusal}, which weights the two means equally and gives one direction for all inputs. Before any token is generated, we rotate the last hidden state of the question's direct prompt once, so that it points along the task-solving direction and keeps its norm.

On \llama{}, these directions raise mean accuracy from $46\%$ under the direct prompt to $85\%$ (CoT: $69\%$), and on \qwen{} from $52\%$ to $86\%$ (CoT: $77\%$). They thus lead to the correct answer more often than the chains they are built from, with 19 generated tokens per question on average, about a tenth of the 197 that \llama{} and the 245 that \qwen{} generate under CoT. The model generates the answer within a few tokens where it can be read from the question, and works through short intermediate steps where it has to be computed. The directions also raise accuracy on tasks where CoT lowers it, such as Movie Recommendation and Ruin Names. They also recover the correct answer on most questions that the CoT and direct prompts both answer incorrectly.

An average of these directions over a task does not give the same result. It leaves mean accuracy at the level of the direct prompt, even on the questions it is built from, so each direction is specific to its own question. Overall, our findings show that for most questions, the representation space of the direct prompt holds a direction that leads the model to the correct answer, and that generating the chain is not the shortest route to it. The question--direction pairs we collect form a dataset which can be used for training a model to predict the direction from the direct prompt, and thereby to match or exceed CoT's accuracy with a fraction of its tokens; we leave this to future work.

\section{Related Work}
\label{sec:related}

\paragraph{What hidden states encode.}
The hidden states of a model often hold more than its output shows. Models can predict whether they know an answer \citep{kadavath2022language}, and probes on their hidden states may reveal the correct answer while they generate a wrong one \citep{orgad2025llms}, as well as whether a CoT will succeed \citep{afzal2025knowing} and whether an intermediate answer is correct \citep{zhang2025reasoning}. Activation patching offers another way to inspect hidden states, by transferring a state into another run and observing how the model's output changes \citep{meng2022locating,zhang2024towards,ghandeharioun2024patchscopes}. Applied to CoT, it shows that the hidden states of a chain can recover the correct answer even when the chain ends in a wrong answer \citep{whencotfails}. We extend this patching to 25 tasks, use its outcomes as labels to probe for shared structure, and intervene with directions built from them.

\paragraph{Directions in the representation space.}
Many concepts and behaviours are represented linearly, and a direction for one can be found as the difference between the mean activations on two contrasting sets of inputs \citep{marks2024geometry,arditi2024refusal,rimsky-etal-2024-steering}. Adding a direction to the activations steers generation \citep{turner2024steering,li2023inference}. For reasoning, activation vectors elicit step-by-step reasoning without a CoT prompt \citep{zhang2025uncovering}, raise accuracy \citep{hojer2025improving} or shorten the chain \citep{chen2025seal,azizi2025activation}. Each of these vectors contrasts two sets of prompts or generations, such as CoT and direct prompts \citep{zhang2025uncovering} or reflection and execution thoughts \citep{chen2025seal}, and a single vector serves every input. Ours instead contrast two sets of states from a question's own chain, split by their patch outcome, so each question has its own direction. Applied to the direct prompt, these directions lead to the correct answer more often than the chains they are built from with far fewer tokens, and the model writes out steps only where the question requires them.

\paragraph{Reasoning with fewer tokens.}
Training models to internalise the chain \citep{deng2024explicit}, to reason in a continuous latent space \citep{hao2025training} or to compute over pause tokens \citep{goyal2024think} lets them reason without generating a chain. We instead keep the model's parameters fixed and show that rotating one hidden state per question makes the direct prompt reason with far fewer tokens. 

\section{Models, Datasets \& Patching}
\label{sec:setup}

\subsection{Models and Datasets}

We use two open-weight instruction-tuned models, \llama{} \citep{llama3} and \qwen{} \citep{qwen}, and evaluate their CoT and direct performance on the 25 tasks.

We categorise the tasks into six reasoning types inspired by the categories of \citet{kazemi-etal-2025-big}.\footnote{We categorise \textit{Penguins in a Table} as logical since its answers follow a logical deduction.} Table~\ref{tab:baselines} shows the CoT and direct baselines. CoT raises mean accuracy by 23 points on \llama{} and 25 on \qwen{} with 197 and 245 tokens per question, but lowers it on Movie Recommendation and Ruin Names for both models and on Causal Judgement for \qwen{}.

\subimport{Tables/}{baselines}

\begin{figure*}[t]
  \centering
  \includegraphics[width=\textwidth]{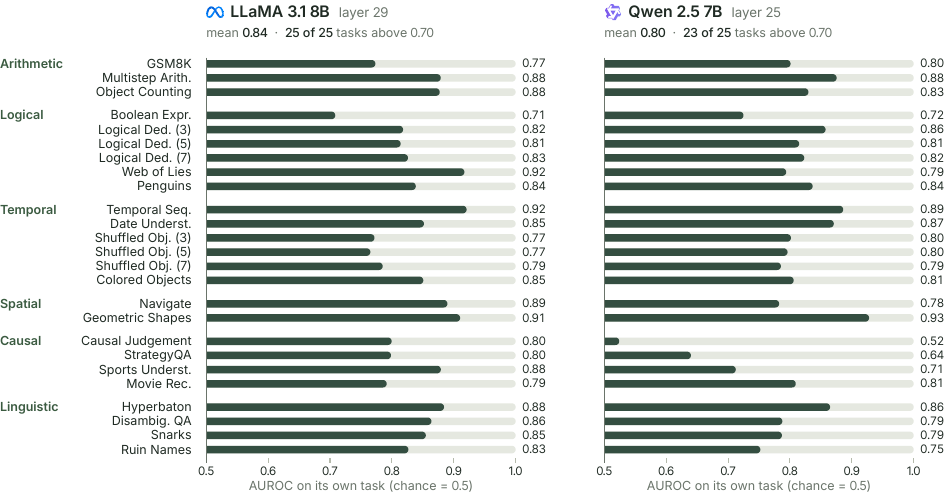}
  \caption{Out-of-fold AUROC of each task's probe, at layer~29 of \llama{} and layer~25 of \qwen{}.}
  \label{fig:probe-auroc}
\end{figure*}

\subsection{Patching Setup}
\label{sec:patching}

We follow the patching setup of \citet{whencotfails}. For each question, we run the model on two prompts that differ only in their system instruction (Table~\ref{tab:prompts} in Appendix~\ref{sec:implementation}). The \emph{source} prompt asks for step-by-step reasoning, and from it we cache the hidden state $h_i^{\ell}$, the output of transformer block $\ell$ at position $i$, for every generated CoT token $i = 1, \dots, n$.\footnote{Patching cost grows with the number of cached states, so we cache three late layers, the \emph{late band}: 27, 29 and 31 for \llama{} and 23, 25 and 27 for \qwen{}.} The \emph{target} prompt asks for just the answer and nothing else, and we write $z^{\ell}$ for its hidden state at layer $\ell$ at the final prompt position, after which the first answer token is generated.

A patch replaces this target state with a cached source state from the same layer, $z^{\ell} \leftarrow h_i^{\ell}$, after which the forward pass continues through the layers above $\ell$ and the model generates as usual.
The model and the target prompt are unchanged and the decoding is greedy, so any change in the answer comes from the patched state. We extract the final answer from the generation and label the state $y_i^{\ell} = 1$ if it matches the gold answer and $y_i^{\ell} = 0$ otherwise. States with $y_i^{\ell} = 1$ are the recovering states and states with $y_i^{\ell} = 0$ the failed states. 
We label the states of every question, including those whose chain ends in a wrong answer, since a state from such a chain can still recover the correct one.

\section{Probing Recovering and Failed States}
\label{sec:probing}

\subsection{Methodology}
\label{sec:probe-method}

We build linear probes, which are simple enough to read what a state encodes rather than learn the task themselves \citep{hewitt-liang-2019-designing,belinkov2022probing}, and ask whether the patch label can be decoded from a single hidden state. Unlike \citet{zhang2025reasoning}, who probe reasoning traces for the correctness of the model's own answer, our target is the outcome $y_i^{\ell}$ of patching the state into the direct run.

The probe takes \emph{CoT states} as inputs, the hidden states $h_i^{\ell}$ of a question's CoT tokens, and we fit one probe for every task and every layer of the late band. Each probe maps a CoT state to the probability that patching it recovers the answer,
\begin{equation}
  p_i^{\ell} = \operatorname{sigmoid}\bigl(w^{\top} \tilde{h}_i^{\ell} + b\bigr),
  \label{eq:probe}
\end{equation}
where $\tilde{h}_i^{\ell}$ is the standardised hidden state. The probe minimises an $\ell_2$-penalised binary cross-entropy on a class-balanced training set (Appendix~\ref{app:probe-training}).

Since the states of one generated reasoning chain are strongly correlated, we divide each task's questions into five folds and score each fold with a probe trained on the other four, so no state is evaluated by a probe that has seen its chain. Since most patches fail (on average $17\%$ of CoT states recover the answer for \llama{} and $28\%$ for \qwen{}), we report the area under the ROC curve (AUROC). Unless stated otherwise we report the middle layer of the late band, layer 29 for \llama{} and layer 25 for \qwen{}, which is also where we do the intervention (Section~\ref{sec:directions}); Appendix~\ref{app:probe-layers} reports all three layers.

\begin{figure*}[t]
  \centering
  \includegraphics[width=\textwidth]{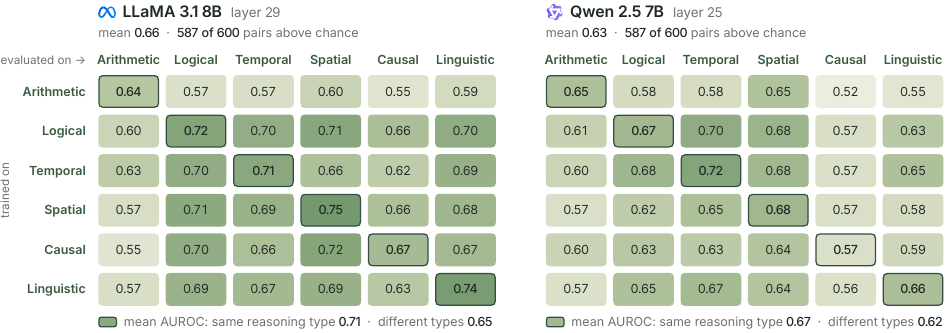}
  \caption{Transfer of the probes across reasoning types, at layer~29 of \llama{} and layer~25 of \qwen{}. Each cell is the mean AUROC of probes trained on a task of the row's type and evaluated on a different task of the column's type; outlined cells pair tasks of the same type. Under each matrix is the mean AUROC over all pairs of tasks of the same type and over all pairs of tasks of different types.}
  \label{fig:probe-transfer}
\end{figure*}

\subsection{Separating Recovering and Failed States}
\label{sec:decodability}

Figure~\ref{fig:probe-auroc} reports the out-of-fold AUROC of each task's probe. The patch label is linearly decodable from single CoT states in every task for \llama{} and in all but one task for \qwen{} (see Appendix~\ref{app:probe-controls} for control tasks). Averaged over the 25 tasks, the AUROC is $0.84$ for \llama{} (range $0.71$--$0.92$) and $0.80$ for \qwen{} (range $0.52$--$0.93$). The weakest probe is on Causal Judgement for \qwen{} ($0.52$). One possible explanation is that the two classes are less easily distinguishable by a linear probe on this task than on others, mostly because of how \qwen{} handles and encodes them. In terms of the percentage of recovering states, this task is similar to other binary tasks such as Navigate and Boolean Expressions (Appendix~\ref{app:dataset}), whose probes are much stronger. Yet the directions of Section~\ref{sec:directions}, built from these states, raise the accuracy here to $86\%$, above CoT.

How well the patch label can be decoded does not depend on how accurate the model is on the task. Across tasks, probe AUROC is not significantly correlated with the model's CoT accuracy, its direct accuracy or the gain of CoT over the direct prompt (Spearman's $|\rho| \le 0.31$, $p > 0.13$; Appendix~\ref{app:probe-per-task}), so the probe is not strong only on the tasks the model already answers well. The label is decodable even where the direct prompt almost never produces the answer: \llama{} answers Multistep Arithmetic correctly in $0.4\%$ of cases without CoT, yet its probe reaches an AUROC of $0.88$.

\subsection{Generalising to Unseen Tasks}
\label{sec:probe-transfer}

To test this on unseen tasks, we refit each task's probe on all of its questions and evaluate it on every CoT state of every other task. Figure~\ref{fig:probe-transfer} averages the resulting $25 \times 25$ matrices over reasoning types. For both models, a probe evaluated on another task scores above chance in 587 of the 600 pairs of tasks (Figure~\ref{fig:probe-transfer-full}, Appendix~\ref{app:probe-transfer}).

This shows that the separation is partly shared between reasoning tasks. For \llama{}, transfer among logical, temporal, spatial, causal and linguistic tasks ($0.62$ to $0.75$) is stronger than transfer from arithmetic to the other types ($0.55$ to $0.60$) or from them to arithmetic ($0.55$ to $0.63$). A potential reason is that arithmetic tasks are harder and require multi-step computation, so the structure their recovering states share may not extend to the other tasks. \qwen{} shows a similar pattern for arithmetic. Transfer to causal tasks is also low ($0.52$ to $0.57$, even from other causal tasks), as three of them are \qwen{}'s hardest targets, including Causal Judgement, on which its own probe is weakest (Section~\ref{sec:decodability}).

Transfer is strongest between tasks of the same reasoning type: pairs of tasks of the same type reach a mean AUROC of $0.71$ for \llama{} and $0.67$ for \qwen{}, against $0.65$ and $0.62$ for pairs of different types (Appendix~\ref{app:probe-transfer}). Recovering and failed states thus differ systematically, within a task and partly across tasks. Section~\ref{sec:directions} builds a direction for each question from this difference and applies it to the direct prompt.

\section{Finding Task-Solving Directions}
\label{sec:directions}

\subsection{Methodology}
\label{sec:direction-method}

\paragraph{Recovering and failed means.}
For each question $q$, we average its recovering states and its failed states,
\begin{equation}
  c^{+}_{q} = \frac{1}{|\mathcal{R}_q|} \sum_{i \in \mathcal{R}_q} h_i^{\ell},
  \qquad
  c^{-}_{q} = \frac{1}{|\mathcal{F}_q|} \sum_{i \in \mathcal{F}_q} h_i^{\ell},
  \label{eq:means}
\end{equation}
where $\mathcal{R}_q = \{i : y_i^{\ell} = 1\}$ and $\mathcal{F}_q = \{i : y_i^{\ell} = 0\}$ index the recovering and failed states of the CoT of $q$ at layer~$\ell$ (Section~\ref{sec:patching}). We call $c^{+}_{q}$ the \emph{recovering mean} of $q$ and $c^{-}_{q}$ its \emph{failed mean}. The states of each set come from different tokens of the chain, so the content of individual tokens partly cancels in the average, while what the states of the set have in common remains. The failed mean is built from the same chain and layer as the recovering mean and differs only in the label of the states it averages, which makes it a matched control for the recovering mean. We build directions from these means rather than from the weights of the probes of Section~\ref{sec:probing}: the weights of a linear classifier also cancel noise and do not, in general, point along the difference between the classes \citep{haufe2014interpretation}, as Appendix~\ref{app:probe-directions} shows for our probes.

\paragraph{Contrastive direction.}
Since the recovering and failed means sit close together,\footnote{Averaged over the 25 tasks, the two means of a question sit about $27^\circ$ apart for \llama{} (cosine $0.89$) and $21^\circ$ for \qwen{} (cosine $0.94$), closer than the recovering means of two different questions of the same task (cosine $0.83$ for \llama{} and $0.90$ for \qwen{}).} a direction close to one is also close to the other. We therefore seek a direction close to the recovering mean and far from the failed mean: the unit vector $x$ that maximises $\langle x, c^{+}_{q} \rangle - \lambda \langle x, c^{-}_{q} \rangle$, with $\lambda \ge 0$ weighting the penalty, lies along $c^{+}_{q} - \lambda\, c^{-}_{q}$. We rescale it to the norm of the recovering mean, so that $\lambda$ changes only the direction,
\begin{equation}
  d_q(\lambda) = \|c^{+}_{q}\| \,
  \frac{c^{+}_{q} - \lambda\, c^{-}_{q}}{\|c^{+}_{q} - \lambda\, c^{-}_{q}\|},
  \label{eq:direction}
\end{equation}
and call $d_q(\lambda)$ the \emph{contrastive direction}. At $\lambda = 0$ it is the recovering mean, and at $\lambda = 1$ it points along the difference-in-means direction used to find and steer concepts and behaviours \citep{marks2024geometry, rimsky-etal-2024-steering, arditi2024refusal}.

\paragraph{Task-level directions.}
To test whether the questions of a task share one direction, the \emph{task-level direction} $d_T(\lambda)$ of a task $T$ averages $d_q(\lambda)$ over the questions of $T$ that have a direction.

\paragraph{Spherical intervention.}
We intervene where the patch labels are defined, on the hidden state $z^{\ell}$ of the direct prompt at its last position, and keep its norm: following \citet{you2026spherical}, we rotate $z^{\ell}$ towards a target $v$, one of the directions above, with Spherical Linear Interpolation \citep[SLERP;][]{shoemake1985animating},
\begin{equation}
  z^{\ell} \leftarrow \|z^{\ell}\| \,
  \frac{\sin\bigl((1-t)\,\theta\bigr)\, \hat{z} + \sin(t\,\theta)\, \hat{v}}{\sin\theta},
  \label{eq:slerp}
\end{equation}
where $\hat{z}$ and $\hat{v}$ are the unit vectors along the original state and the target, $\theta$ is the angle between them, and $t$ is the fraction of this angle through which the original state is rotated. At $t = 1$, which we use unless stated otherwise, the state points along the target (Appendix~\ref{app:directions-robustness} varies $t$, the norm and the layer). The rotation is applied once, after which the model generates as usual.

Additive steering \citep{turner2024steering, rimsky-etal-2024-steering} and patching change the norm of the state along with its direction. The rotation changes the direction alone, which keeps generation intact where additive steering can degrade it \citep{you2026spherical, dang2026selective}. At $t = 1$ it is a norm-matched patch: the state keeps the norm the direct prompt produced and takes only its direction from the chain, so any gain is carried by a direction in the representation space of the direct prompt.

\paragraph{Evaluation.}
For every question, we rotate the state towards its recovering mean, its failed mean, $d_q(\lambda)$ with $\lambda \in \{0.25, 0.5, 0.75\}$ and the task-level direction $d_T(0.5)$ of its task, and compare accuracy and generated tokens with the direct prompt and with the CoT from which the directions are built.

\begin{table}[t]
\centering
\footnotesize
\renewcommand{\arraystretch}{1.06}
\setlength{\tabmodelgap}{0.8em}%
\begin{tabular*}{\columnwidth}{@{\extracolsep{\fill}}>{\hspace{0.45em}}l
    @{\hspace{0.2em}}r@{\hspace{0.35em}}r@{\hspace{0.35em}}r@{\hspace{0.55em}}r
    @{\hspace{\tabmodelgap}}r@{\hspace{0.35em}}r@{\hspace{0.35em}}r@{\hspace{0.55em}}r@{}}
\toprule
 & \multicolumn{4}{@{}c@{\hspace{\tabmodelgap}}}{\modelwithlogo{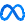}{\llama}}
 & \multicolumn{4}{@{}c@{}}{\modelwithlogo{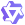}{\qwen}} \\
\cmidrule(r{\tabmodelgap}){2-5}\cmidrule{6-9}
 & \colhead{Acc.} & \colhead{$\Delta$\textsubscript{Dir}} & \colhead{$\Delta$\textsubscript{CoT}} & \colhead{Tok.} & \colhead{Acc.} & \colhead{$\Delta$\textsubscript{Dir}} & \colhead{$\Delta$\textsubscript{CoT}} & \colhead{Tok.} \\
\midrule
\addlinespace[0.25em]
\taskgroup{Prompt} \\[-1.2pt]
CoT & 69 & \dpos{23} & -- & 197 & 77 & \dpos{25} & -- & 245 \\
Direct & 46 & -- & \dneg{23} & 1 & 52 & -- & \dneg{25} & 1 \\
\groupskip
\taskgroup{Question-specific rotation} \\[-1.2pt]
Recovering $c^{+}_{q}$ & 80 & \dpos{34} & \dpos{11} & 17 & 82 & \dpos{30} & \dpos{5} & 18 \\
$d_{q}(0.25)$ & 83 & \dpos{38} & \dpos{15} & 18 & 85 & \dpos{33} & \dpos{8} & 19 \\
$d_{q}(0.5)$ & \textbf{85} & \dpos{39} & \dpos{16} & 19 & \textbf{86} & \dpos{34} & \dpos{10} & 19 \\
$d_{q}(0.75)$ & 83 & \dpos{37} & \dpos{14} & 18 & 84 & \dpos{32} & \dpos{7} & 19 \\
Failed $c^{-}_{q}$ & 9 & \dneg{37} & \dneg{60} & 4 & 16 & \dneg{36} & \dneg{61} & 6 \\
\groupskip
\taskgroup{Task-level rotation} \\[-1.2pt]
$d_{T}(0.5)$ & 46 & 0 & \dneg{23} & 8 & 50 & \dneg{2} & \dneg{27} & 15 \\
\addlinespace[0.25em]
\bottomrule
\end{tabular*}
\caption{Rotating the last state of the direct prompt, at layer~29 of \llama{} and layer~25 of \qwen{} ($t = 1$). Acc.\ (\%) is macro-averaged over the 25 tasks, counting a question without a direction as incorrect; $\Delta$s, computed before rounding, are from the direct and CoT prompts, and Tok.\ is the mean number of generated tokens. Appendix~\ref{app:directions-per-task} gives every task.}
\label{tab:interventions}
\end{table}

\begin{figure*}[t]
  \centering
  \includegraphics[width=\textwidth]{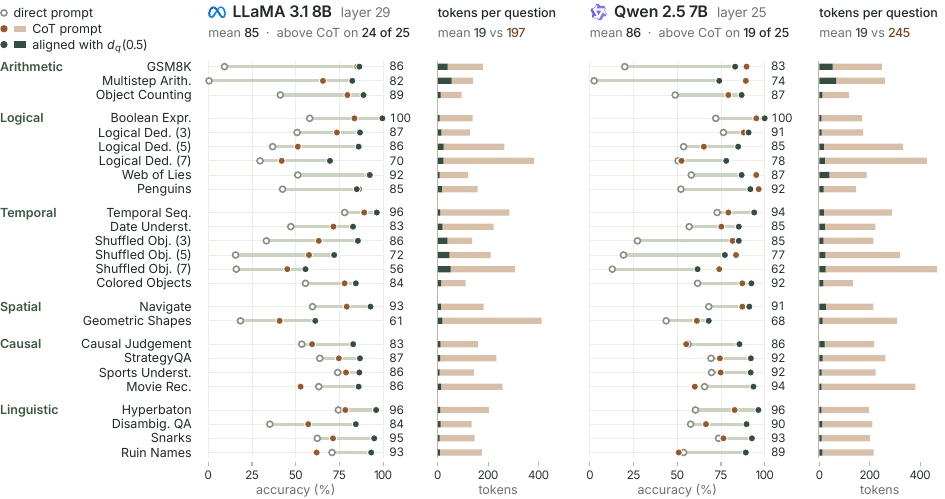}
  \caption{Accuracy and generated tokens per task when the state of the direct prompt is aligned with $d_q(0.5)$ (green), against the direct (grey ring) and CoT (brown) prompts, at layer~29 of \llama{} and layer~25 of \qwen{} ($t = 1$). The number at the end of each row is the accuracy with $d_q(0.5)$. Each token bar runs to the tokens CoT generates per question, and its green part to those generated with $d_q(0.5)$; the direct prompt generates about one.}
  \label{fig:question-tasks}
\end{figure*}

\subsection{Question-Specific Directions}
\label{sec:question-directions}

Table~\ref{tab:interventions} reports the intervention results. Aligning the hidden state of the direct prompt with the recovering mean of its question raises accuracy from 46 to 80 on \llama{} and from 52 to 82 on \qwen{}, above the 69 and 77 of CoT, with 17 and 18 generated tokens per question against 197 and 245. Aligning it with the failed mean, built from the same chain, lowers accuracy to 9 and 16, far below the direct prompt. This indicates that in the representation space of a model instructed to answer directly, there exists a direction for most questions that, if the state is aligned with it, leads the model to much shorter generations that reach the correct answer more often than CoT does.

The contrastive direction improves the performance even further. Accuracy rises with $\lambda$ to 85 on \llama{} and 86 on \qwen{} at $\lambda = 0.5$, and falls again at $\lambda = 0.75$. Increasing $\lambda$ moves the target away from the failed mean, which raises accuracy, but also away from the nearby recovering mean, which lowers it; up to $\lambda = 0.5$ the gain outweighs the loss. Using both means therefore helps find a better target than the recovering mean alone. At $\lambda = 0.5$ the direction improves on the direct prompt by 39 and 34 points and on CoT by 16 and 10, with 19 generated tokens per question for both models. Among the questions that both prompts answer incorrectly, $d_q(0.5)$ recovers the correct answer on $64\%$ for \llama{} and $70\%$ for \qwen{} (Appendix~\ref{app:examples}), even though it is built from a chain that ends in a wrong answer.

\begin{figure*}[t]
  \centering
  \includegraphics[width=\textwidth]{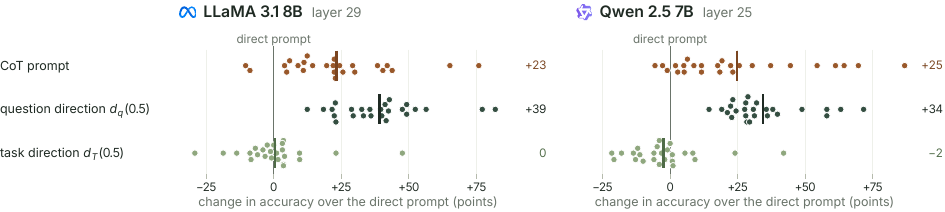}
  \caption{Change in each task's accuracy over the direct prompt (zero line) under CoT and when the state is aligned with each question's own direction $d_q(0.5)$ or with the task-level direction $d_T(0.5)$, at layer~29 of \llama{} and layer~25 of \qwen{} ($t = 1$). Dark ticks give the row means.}
  \label{fig:task-level}
\end{figure*}

Figure~\ref{fig:question-tasks} breaks the result down by task. Aligning the state of the direct prompt with $d_q(0.5)$ improves on the direct prompt on every task for both models, and on the CoT prompt on 24 of the 25 tasks for \llama{} and 19 for \qwen{}. Of the six tasks where it stays below CoT for \qwen{}, three are Penguins in a Table, Web of Lies and GSM8K, on which CoT is already strong: \qwen{}'s CoT reaches 90 to 97, and for \llama{} the direction only matches CoT. On the other three, Multistep Arithmetic and Tracking Shuffled Objects with five and seven objects, a potential reason is that \qwen{}'s chains have few recovering states on these tasks: of all 25 tasks, these three have the smallest share (Table~\ref{tab:dataset} in Appendix~\ref{app:dataset}). The direction achieves comparable accuracy on these three tasks for both models and is above CoT for \llama{}, whose CoT accuracy on them is much lower.

On the tasks where CoT fails to improve over the direct prompt, Movie Recommendation and Ruin Names for both models and Causal Judgement for \qwen{}, the direction still raises accuracy, by 22 to 36 points over the direct prompt. A plausible explanation is that the first two need no intermediate steps that build on each other. They ask the model to identify an answer, namely a film similar to four given ones or a humorous edit of a name. The chain therefore adds no computation, and its steps may instead confuse the model and sway it away from its original answer, much as deliberation can hurt recognition in humans \citep{liu2025mind}.

With the direction, the model works through intermediate steps only where a task needs them. Where the answer can be read from the question, the model answers within a few tokens. \llama{} generates 3 tokens per question on Boolean Expressions and 5 on Snarks, and \qwen{} generates 2 and 9. GSM8K and Multistep Arithmetic require the answer to be computed, and here the model writes out the steps without the extra prose. \llama{} generates 39 and 55 tokens on these tasks, against 179 and 140 under CoT, and \qwen{} generates 52 and 66, against 248 and 260. On a GSM8K question, \llama{} answers 121 under the direct prompt and generates ``\ldots{} $\$21 + \$50 = \$71$'' with the direction (Appendix~\ref{app:examples}). The direction thus sets the model on the computation the question needs.

A prompt is chosen once for a whole task, yet the direct prompt fails when the answer has to be computed and CoT lowers accuracy on tasks that need no intermediate steps. The representation space of the direct prompt holds a direction for each question that avoids both failures. Pointed towards these directions, the model reaches a mean accuracy of 85 on \llama{} and 86 on \qwen{}, while generating about a tenth of the tokens.

\subsection{Task-Level Directions}
\label{sec:task-directions}

Each question-specific direction is built from the generated CoT and patch labels of its own question, and we use it to establish what the representation space of the direct prompt can reach. If the questions of a task shared a direction, it could be built once and applied to new questions of the task without generating their chains or patching their states.

We test whether such a direction exists with the task-level direction $d_T(0.5)$ applied to the questions of its task. Since it is built from these same questions, this is the most favourable case for a shared direction. Even so, it leaves accuracy at the level of the direct prompt (Figure~\ref{fig:task-level}): 46 against 46 on \llama{} and 50 against 52 on \qwen{}, while the questions' own directions reach 85 and 86 on the same questions.

These results indicate that the task-solving direction is specific to each question. A direction averaged over the questions of a task rarely improves on the direct prompt by more than a few points, let alone reaches the accuracy of CoT. At the level of individual questions, however, the representation space of the direct prompt contains such a direction for most questions. Aligning the state with it yields the correct answer with about a tenth of CoT's generated tokens, and intermediate steps appear only where the answer must be computed. The resulting question--direction pairs constitute a dataset for training a model to predict the direction from the direct prompt, and to match or exceed the accuracy of CoT while generating far fewer tokens; we leave this to future work.

\section{Conclusion}
\label{sec:conclusion}

In this paper, we show that a linear probe can distinguish the CoT hidden states that are useful for recovering the correct answer to a question from those that are not, across 25 tasks spanning six reasoning types. We further show that this distinction can be used to build, for each question, a direction that reduces the discrepancy between what the model generates in CoT and what it computes internally. Finally, we find that this direction is specific to each question and is not shared between different questions of the same task or between different tasks. Together, these results show that, for the large majority of questions, there exists a direction in the representation space of the direct prompt that makes the model answer correctly while generating about a tenth as many tokens as CoT, so CoT is not the most efficient route to the answer. The question--direction pairs we obtain can be used as a dataset to train models to align their representations with these directions, which we leave to future work.

\section*{Limitations}

Each direction is built from the patch labels of its question's own chain, and these labels compare the answer after patching with the gold answer. Producing the direction of a new question without its chain and labels, for example with a model trained on the directions we collect, is left to future work. A first attempt, a regressor that predicts the direction from the direct prompt alone, leaves accuracy at the level of the direct prompt. We study two instruction-tuned models of 7 and 8 billion parameters, and larger models or models trained to reason at length may encode recovering and failed states differently. Our tasks have short answers that can be checked automatically, and we excluded three BIG-Bench Hard tasks on which answer extraction after patching was unreliable and one on which the models often produced no chain. We cache and intervene at three late layers and at the last position of the direct prompt only, the position found most effective for patching \citep{whencotfails}.

\section*{Ethical Considerations}

We use all external artifacts for research only and in line with their licences: GSM8K, StrategyQA and BIG-Bench Hard under the MIT licence, \llama{} under the Llama 3.1 Community License and its Acceptable Use Policy, and \qwen{} under the Apache 2.0 licence. The datasets are public benchmarks of reasoning questions. We analyse and intervene on the hidden states of these models to understand how they solve tasks; we release no new model, and our interventions are not intended for deployment. The artifacts we derive consist of experimental results, analyses and figures.

\bibliography{custom}

\appendix
\raggedbottom 

\section{Implementation Details}
\label{sec:implementation}
The experiments use \llama{}\footnote{\url{https://huggingface.co/meta-llama/Llama-3.1-8B-Instruct}} and \qwen\footnote{\url{https://huggingface.co/Qwen/Qwen2.5-7B-Instruct}} in their instruction-tuned variants and are run on NVIDIA GH200 nodes, using about 8{,}000 GPU hours in total, including exploratory and preliminary experiments. All experiments are run with a batch size of $1$ and the seed $42$ to ensure reproducibility. We implement them in PyTorch \citep{paszke2019pytorch} with the Hugging Face Transformers library \citep{wolf-etal-2020-transformers}, and compute AUROC with scikit-learn \citep{pedregosa2011scikit} and the statistical tests with SciPy \citep{virtanen2020scipy}.

\paragraph{Prompts.}
Table~\ref{tab:prompts} gives the prompt templates: both prompts of a question share the user turn and differ only in the system instruction, whose answer slot depends on the answer format of the task. Both prompts are formatted with the model's chat template, and $z^{\ell}$ is the state of the last token of the formatted prompt.

\paragraph{Data.}
We use the test split of GSM8K (1{,}319 questions), the labelled set of StrategyQA (2{,}290 questions) and all questions of 23 of the 27 BIG-Bench Hard subtasks (Table~\ref{tab:dataset}). We exclude \textit{Dyck Languages}, \textit{Word Sorting} and \textit{Formal Fallacies}, where answer extraction after patching was not sufficiently clean-cut for both models, and \textit{Salient Translation Error Detection}, where the models often failed to generate a CoT trace. By answer format, three tasks are numeric (GSM8K, Multistep Arithmetic and Object Counting), six are binary (StrategyQA, Boolean Expressions, Causal Judgement, Navigate, Sports Understanding and Web of Lies), and the other 16 subtasks are multiple choice.

\subimport{Tables/}{prompts}

\paragraph{Decoding.}
Decoding is greedy in every run. The CoT and direct baselines (Table~\ref{tab:baselines}) generate at most 1{,}024 tokens. For patching, the CoT run from which we cache states and every patched run generate at most 400 tokens, and every run with an intervention (Section~\ref{sec:directions}) at most 128. The CoT run for patching reaches its limit on 511 of the 9{,}120 questions for \llama{} and 693 for \qwen{}, and the CoT baseline reaches its own on 123 questions for \llama{} and 1 for \qwen{}. Generated tokens are counted with the model's tokenizer, excluding special tokens.

\paragraph{Answer extraction.}
We evaluate the models with a flexible answer extraction, similar to the answer cleansing of \citet{kojima2022large}, and we apply it to every generation: under the CoT and direct prompts, after a patch and after an intervention. We read the answer from the text after the last \texttt{Final Answer:} of the generation, and from the whole generation if it has no such marker or no answer after it. Only the CoT prompt asks for this marker, so the other generations are almost always read whole ($99.7\%$ of the patched runs of both models), as are 595 of the 9{,}120 chains of \llama{}'s CoT run for patching and 667 of \qwen{}'s, most of which reach the token limit. What we read depends on the answer format of the task: the last number for numeric tasks, with thousands separators and trailing zeros removed; the last \textit{True}, \textit{False}, \textit{Yes} or \textit{No} for binary tasks; and the last standalone capital letter, with or without parentheses, for multiple-choice tasks. The answer is correct if it matches the gold answer, ignoring case; for the multiple-choice tasks of BIG-Bench Hard, we remove the parentheses from the gold label, so that (A) becomes A. A generation from which no answer can be read counts as incorrect, so a patch whose continuation states no answer is a failed patch.

\section{Patch Labels}
\label{app:dataset}

Table~\ref{tab:dataset} describes the labelled states of each task. Most patches fail. Averaged over tasks, $17\%$ of the CoT states of \llama{} recover the answer, from $4\%$ on Geometric Shapes to $47\%$ on Boolean Expressions, and $28\%$ of those of \qwen{}, from $7\%$ on Tracking Shuffled Objects (7) to $56\%$ on Boolean Expressions. On the binary tasks of \qwen{} the share approaches one half. Almost every question has both a recovering and a failed state, and so a direction; the exceptions concentrate in the hardest tasks, Tracking Shuffled Objects (7), Geometric Shapes and Logical Deduction (7). These exceptions, 350 questions for \llama{} and 192 for \qwen{}, all lack a recovering state, as no question has only recovering states, and we count them as incorrect under every direction without running them.

\subimport{Tables/}{dataset}


\begin{table*}[t]
  \centering
  \small
  \setlength{\tabcolsep}{4pt}
  \begin{tabular}{lrcccccc}
    \toprule
    & & \multicolumn{3}{c}{\modelwithlogo{icons/meta-color.pdf}{\llama{}}, layer 29} & \multicolumn{3}{c}{\modelwithlogo{icons/qwen-color.pdf}{\qwen{}}, layer 25} \\
    \cmidrule(lr){3-5} \cmidrule(lr){6-8}
    Task & $n$ & AUROC & Sel.\ orig. & Sel.\ w.-pos. & AUROC & Sel.\ orig. & Sel.\ w.-pos. \\
    \midrule
    GSM8K & 1{,}319 & 0.773 {\scriptsize$\pm$0.007} & 0.219 & 0.199 & 0.801 {\scriptsize$\pm$0.008} & 0.262 & 0.209 \\
    Multistep Arith. & 250 & 0.879 {\scriptsize$\pm$0.012} & 0.132 & 0.180 & 0.875 {\scriptsize$\pm$0.029} & 0.288 & 0.294 \\
    Object Counting & 250 & 0.877 {\scriptsize$\pm$0.015} & 0.235 & 0.346 & 0.830 {\scriptsize$\pm$0.015} & 0.209 & 0.272 \\
    Boolean Expr. & 250 & 0.709 {\scriptsize$\pm$0.032} & 0.100 & 0.201 & 0.725 {\scriptsize$\pm$0.006} & 0.122 & 0.214 \\
    Logical Ded.\ (3) & 250 & 0.818 {\scriptsize$\pm$0.018} & 0.139 & 0.254 & 0.858 {\scriptsize$\pm$0.010} & 0.241 & 0.317 \\
    Logical Ded.\ (5) & 250 & 0.814 {\scriptsize$\pm$0.022} & 0.246 & 0.291 & 0.815 {\scriptsize$\pm$0.017} & 0.250 & 0.313 \\
    Logical Ded.\ (7) & 250 & 0.826 {\scriptsize$\pm$0.020} & 0.254 & 0.326 & 0.823 {\scriptsize$\pm$0.011} & 0.270 & 0.312 \\
    Web of Lies & 250 & 0.917 {\scriptsize$\pm$0.009} & 0.221 & 0.297 & 0.794 {\scriptsize$\pm$0.006} & 0.191 & 0.270 \\
    Penguins & 146 & 0.839 {\scriptsize$\pm$0.010} & 0.240 & 0.320 & 0.837 {\scriptsize$\pm$0.013} & 0.274 & 0.294 \\
    Temporal Seq. & 250 & 0.921 {\scriptsize$\pm$0.003} & 0.300 & 0.348 & 0.886 {\scriptsize$\pm$0.005} & 0.273 & 0.337 \\
    Date Underst. & 250 & 0.852 {\scriptsize$\pm$0.009} & 0.286 & 0.320 & 0.871 {\scriptsize$\pm$0.006} & 0.308 & 0.334 \\
    Shuffled Obj.\ (3) & 250 & 0.772 {\scriptsize$\pm$0.009} & 0.107 & 0.212 & 0.802 {\scriptsize$\pm$0.013} & 0.091 & 0.270 \\
    Shuffled Obj.\ (5) & 250 & 0.765 {\scriptsize$\pm$0.004} & 0.146 & 0.228 & 0.796 {\scriptsize$\pm$0.022} & 0.173 & 0.273 \\
    Shuffled Obj.\ (7) & 250 & 0.785 {\scriptsize$\pm$0.023} & 0.211 & 0.260 & 0.785 {\scriptsize$\pm$0.023} & 0.176 & 0.299 \\
    Colored Objects & 250 & 0.851 {\scriptsize$\pm$0.009} & 0.250 & 0.321 & 0.806 {\scriptsize$\pm$0.014} & 0.190 & 0.275 \\
    Navigate & 250 & 0.890 {\scriptsize$\pm$0.019} & 0.289 & 0.344 & 0.782 {\scriptsize$\pm$0.019} & 0.178 & 0.254 \\
    Geometric Shapes & 250 & 0.910 {\scriptsize$\pm$0.011} & 0.305 & 0.361 & 0.928 {\scriptsize$\pm$0.011} & 0.331 & 0.401 \\
    Causal Judgement & 187 & 0.800 {\scriptsize$\pm$0.017} & 0.243 & 0.280 & 0.524 {\scriptsize$\pm$0.053} & $-$0.021 & 0.042 \\
    StrategyQA & 2{,}290 & 0.799 {\scriptsize$\pm$0.005} & 0.254 & 0.282 & 0.640 {\scriptsize$\pm$0.009} & 0.086 & 0.138 \\
    Sports Underst. & 250 & 0.879 {\scriptsize$\pm$0.011} & 0.228 & 0.304 & 0.712 {\scriptsize$\pm$0.009} & 0.128 & 0.197 \\
    Movie Rec. & 250 & 0.792 {\scriptsize$\pm$0.012} & 0.229 & 0.258 & 0.809 {\scriptsize$\pm$0.003} & 0.211 & 0.282 \\
    Hyperbaton & 250 & 0.884 {\scriptsize$\pm$0.020} & 0.249 & 0.285 & 0.865 {\scriptsize$\pm$0.010} & 0.219 & 0.312 \\
    Disambig.\ QA & 250 & 0.864 {\scriptsize$\pm$0.011} & 0.206 & 0.308 & 0.788 {\scriptsize$\pm$0.005} & 0.111 & 0.243 \\
    Snarks & 178 & 0.855 {\scriptsize$\pm$0.009} & 0.205 & 0.283 & 0.787 {\scriptsize$\pm$0.009} & 0.174 & 0.205 \\
    Ruin Names & 250 & 0.827 {\scriptsize$\pm$0.004} & 0.227 & 0.284 & 0.752 {\scriptsize$\pm$0.012} & 0.180 & 0.229 \\
    \midrule
    Macro average & 9{,}120 & 0.836 & 0.221 & 0.284 & 0.796 & 0.197 & 0.263 \\
    \bottomrule
  \end{tabular}
  \caption{Per-task probes at the layers reported in the main text. $n$: questions, all of which are scored out of fold. AUROC $\pm$ the standard deviation of the per-fold AUROC; Figure~\ref{fig:probe-auroc} shows the same AUROC values to two decimals, rounded from the unrounded values. Sel.\ orig./w.-pos.: selectivity against the origin and within-position controls (Appendix~\ref{app:probe-controls}).}
  \label{tab:probe-per-task}
\end{table*}

\section{Probe}
\label{sec:probe-appendix}

\subsection{Training the Probe}
\label{app:probe-training}

\paragraph{Data and folds.}
The questions of each task $T$ are assigned to five folds, and every state of a question falls in the fold of that question. The assignment is a random permutation of the task's questions, seeded with the task's name so that both models use the same folds. Over the 25 tasks there are $9{,}120$ questions per model, and per layer $1.77$M states for \llama{} and $2.24$M for \qwen{}. Every probe is trained on a union of folds and evaluated only on questions outside it.

\paragraph{Class balance.}
Most patches fail, so within each fold we downsample the failed states to a 50/50 ratio. Any union of folds is therefore balanced. Evaluation always uses all states at their natural label distribution.

\paragraph{Objective.}
On a class-balanced training set $\mathcal{S}$, the probe of Eq.~\eqref{eq:probe} minimises the $\ell_2$-penalised binary cross-entropy
\begin{align}
  \mathcal{J} = {}& {-\frac{1}{|\mathcal{S}|}} \sum_{i \in \mathcal{S}} \bigl[ y_i^{\ell} \log p_i^{\ell} + (1 - y_i^{\ell}) \log (1 - p_i^{\ell}) \bigr] \nonumber \\
  &+ \frac{\alpha}{2} \lVert w \rVert_2^2 ,
  \label{eq:probe-loss}
\end{align}
where the bias is not penalised. This is the standard $\ell_2$-regularised linear probe, with inverse regularisation strength $C = 1 / (\alpha |\mathcal{S}|)$.

\paragraph{Standardisation.}
Each probe standardises its states per dimension,
\begin{equation}
  \tilde{h}_i^{\ell} = \bigl(h_i^{\ell} - \mu\bigr) \oslash \sigma,
  \label{eq:standardise}
\end{equation}
where $\oslash$ is elementwise division and $\mu, \sigma \in \mathbb{R}^{m}$, with $m$ the hidden size, are the per-dimension mean and standard deviation of its training set $\mathcal{S}$. The probe applies the same $\mu$ and $\sigma$ to every state it scores. Per-dimension scaling makes the penalty in Eq.~\eqref{eq:probe-loss} treat all dimensions alike and keeps a few high-variance dimensions from dominating the probe.

\paragraph{Optimisation.}
We minimise Eq.~\eqref{eq:probe-loss} with full-batch L-BFGS \citep{liu1989limited} and a strong-Wolfe line search in double precision, and stop when every coordinate of the gradient is at most $10^{-6}$ in magnitude.

\paragraph{Choosing the penalty.}
The penalty $\alpha$ is chosen from $\{10^{-4}, 10^{-3}, \dots, 10^{3}\}$ by nested cross-validation, so no hyperparameter is chosen on the states a probe is evaluated on. For the probe that scores fold $k$, each $\alpha$ is assessed by leave-one-fold-out cross-validation over the remaining four folds: we train on three and compute the AUROC on all states of the fourth. The value with the highest mean AUROC is selected, and the probe is refitted on all four folds at that value.

\paragraph{Out-of-fold scores.}
Each fold is scored by the probe trained without it, and the five sets of scores are pooled into one AUROC per task and layer.

\paragraph{Full refit for transfer.}
The source probe for transfer (Section~\ref{sec:probe-transfer}) is fitted on all questions of its task. Its $\alpha$ is the value with the highest mean out-of-fold AUROC, each fold being evaluated by the probe trained on the other four. This selection is not nested, but the full refit is only ever evaluated on other tasks.

\subsection{Control Tasks}
\label{app:probe-controls}

A probe can score well without reading anything specific to a state. Each probe sees a single task and layer, but patch success is not uniform along the chain: earlier positions recover more often than later ones \citep{whencotfails}. A probe could therefore reach a high AUROC by inferring a state's position and predicting the success rate of that position. Following \citet{hewitt-liang-2019-designing}, we train control probes with the same architecture on modified labels and report the selectivity of each probe against them, its AUROC minus that of the control probe (Tables~\ref{tab:probe-per-task} and~\ref{tab:probe-layers}). Since the exploitable structure here is a state's position, we assign control labels per position, separately for each task and layer.

The \emph{origin control} gives all states at a position one fixed random label, drawn so that the success rate of the task and layer is preserved. A probe can solve it only by recovering where in the chain a state comes from.

The \emph{within-position control} shuffles the real labels among the states at a position, independently within each fold. Each position thus keeps its exact success rate, but which question recovers becomes random. A probe can solve it only by predicting the success rate of each position.

Control probes are cross-fitted exactly like the real probes, using the penalty chosen for the real probe of the same fold. Selectivity against the origin control measures what the probe decodes beyond a state's position. Selectivity against the within-position control measures whether the probe separates recovering from failed states at the same position. Within-position selectivity is positive for all 50 probes at the layers of the main text (means $0.284$ for \llama{} and $0.263$ for \qwen{}).

\begin{figure*}[t]
  \centering
  \includegraphics[width=\textwidth]{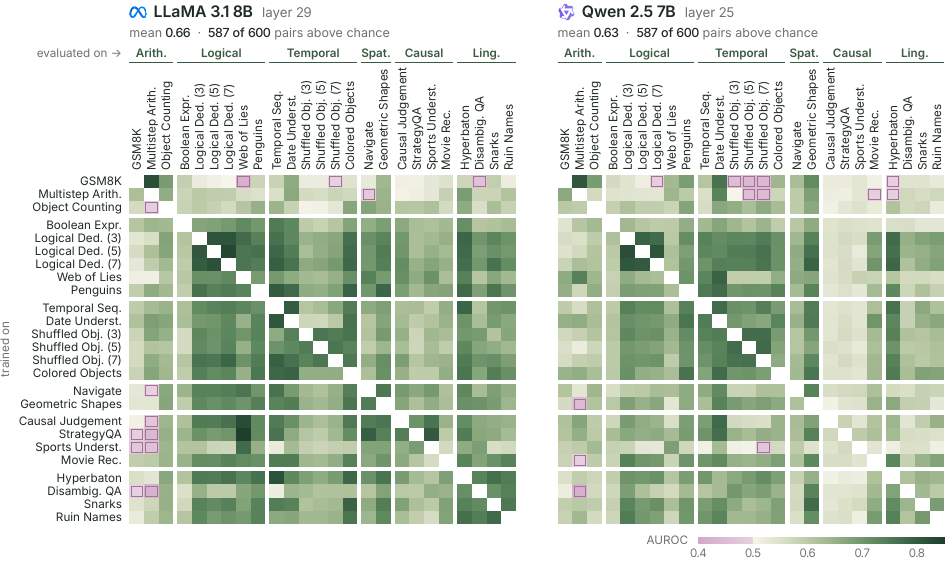}
  \caption{Transfer of the probes between every pair of tasks, at layer~29 of \llama{} and layer~25 of \qwen{}. Each cell is the AUROC of a probe trained on the row's task and evaluated on the column's task; tasks are grouped by reasoning type. Cells run from ivory at chance (0.5) to green, and the diagonal, a probe on its own task (Figure~\ref{fig:probe-auroc}), is left white. Plum cells fall below chance. Figure~\ref{fig:probe-transfer} averages these matrices over reasoning types.}
  \label{fig:probe-transfer-full}
\end{figure*}

\subsection{Per-Task Results}
\label{app:probe-per-task}

Table~\ref{tab:probe-per-task} reports the AUROC and both selectivities of every task's probe at the layers reported in the main text. The estimates are stable across folds: the median standard deviation of the per-fold AUROC is $0.011$. Across the 25 tasks, probe AUROC is not significantly correlated with CoT accuracy (Spearman's $\rho = 0.30$ for \llama{} and $0.16$ for \qwen{}), direct accuracy ($0.31$ and $-0.21$) or the gain of CoT over the direct prompt ($-0.17$ and $0.25$), all with $p > 0.13$. For \qwen{}, probe AUROC is negatively correlated with the share of patches that succeed ($\rho = -0.57$, $p = 0.003$) only because its six binary tasks, on which almost half of the patches succeed ($46\%$, against $22\%$ on the other tasks), have weaker probes on average. Without them, $\rho = -0.14$ ($p = 0.57$). For \llama{} there is no such relation ($\rho = -0.12$, $p = 0.56$).

\subsection{Layer Choice}
\label{app:probe-layers}

Table~\ref{tab:probe-layers} reports all three layers of the late band. For \llama{}, layer~27 is worse than each of layers~29 and~31 on 24 of the 25 tasks ($p < 10^{-6}$ for both AUROC and within-position selectivity; Wilcoxon signed-rank), and layer~31 exceeds layer~29 only slightly ($0.838$ against $0.836$, $p = 0.024$). For \qwen{} no pair of layers differs ($p \ge 0.20$). The layers we report in the main text, layer~29 for \llama{} and layer~25 for \qwen{}, are therefore within $0.003$ AUROC of the best layer of each model.

\subimport{Tables/}{appendix-probe-selectivity-layers}

\begin{table*}[t]
  \centering
  \small
  \begin{tabular}{lcccccc}
    \toprule
    & & \multicolumn{3}{c}{Off-diagonal} & \multicolumn{2}{c}{Reasoning type} \\
    \cmidrule(lr){3-5} \cmidrule(lr){6-7}
    $\ell$ & Diagonal & AUROC & ${>}0.5$ & Sel.\ w.-pos. & Same & Different \\
    \midrule
    \multicolumn{7}{@{}l}{\modelwithlogo{icons/meta-color.pdf}{\llama{}}} \\
    27 & 0.819 & 0.648 & 582/600 & 0.127 & 0.689 & 0.640 \\
    29 & 0.836 & 0.663 & 587/600 & 0.145 & 0.710 & 0.654 \\
    31 & 0.838 & 0.674 & 590/600 & 0.155 & 0.719 & 0.666 \\
    \midrule
    \multicolumn{7}{@{}l}{\modelwithlogo{icons/qwen-color.pdf}{\qwen{}}} \\
    23 & 0.795 & 0.630 & 591/600 & 0.122 & 0.669 & 0.622 \\
    25 & 0.796 & 0.630 & 587/600 & 0.123 & 0.672 & 0.623 \\
    27 & 0.794 & 0.635 & 585/600 & 0.129 & 0.673 & 0.628 \\
    \bottomrule
  \end{tabular}
  \caption{Probe transfer per layer. Diagonal: mean out-of-fold AUROC within each task. Off-diagonal: mean AUROC of a probe trained on all questions of one task and scored on every state of another; ${>}0.5$: cells above chance; Sel.\ w.-pos.: mean selectivity against the within-position control. Same/Different: mean off-diagonal AUROC between tasks of the same or of a different reasoning type.}
  \label{tab:probe-transfer}
\end{table*}

\subsection{Transfer}
\label{app:probe-transfer}

Table~\ref{tab:probe-transfer} summarises the transfer matrices at all three layers, and Figure~\ref{fig:probe-transfer-full} shows them in full at the layers of the main text. For \llama{} the three arithmetic tasks, the only ones with numeric answers, are the three weakest sources, and GSM8K and Multistep Arithmetic the two hardest targets. Yet they transfer better to one another than to or from any other type (Figure~\ref{fig:probe-transfer}), which suggests that the arithmetic tasks share an encoding of recovering and failed states that extends only partly to the other tasks. For \qwen{} the hardest targets are Causal Judgement, Sports Understanding and StrategyQA. All three are binary tasks, and two of them carry \qwen{}'s weakest within-task probes (Figure~\ref{fig:probe-auroc}). Its weakest sources are Sports Understanding, Multistep Arithmetic and GSM8K. Transfer increases with depth for \llama{} and is flat for \qwen{} (Table~\ref{tab:probe-transfer}).

For the Same/Different columns of Table~\ref{tab:probe-transfer} we compare the mean AUROC of the off-diagonal cells between tasks of the same reasoning type (Table~\ref{tab:baselines}) with that of the cells between tasks of different types, 92 against 508 cells. As cells that share a task are not independent, we test the difference by permutation: we shuffle the reasoning types across the 25 tasks, keeping the number of tasks of each type, and none of 100{,}000 shuffles gives a difference as large as the observed one at the layers of the main text ($p < 10^{-5}$ for both models).

In transfer we report only the within-position control (Appendix~\ref{app:probe-controls}). The source task's within-position control probe is scored on the target's within-position labels. The origin control is uninformative off the diagonal, because the source task's arbitrary position labels are unrelated to the target's. The within-position control probes transfer at an AUROC of only $0.518$ (\llama{}) and $0.507$ (\qwen{}), so off-diagonal selectivity is $0.145$ and $0.123$ on average (Table~\ref{tab:probe-transfer}), and positive in $581$ and $587$ of the $600$ cells.

\subsection{Probe Weights and Activation Patterns}
\label{app:probe-directions}

In this section we show why we build directions from the means of recovering and failed states rather than from the probe weights. On the unstandardised state $h$ (Eq.~\eqref{eq:standardise}), the probe's \emph{weight vector} is $w_{\mathrm{raw}} = w \oslash \sigma$. Because the probe is trained to classify, it can weight some dimensions only to cancel noise they share with informative ones, so its weight vector does not necessarily point along the difference between recovering and failed states \citep{haufe2014interpretation}. That difference is described by the probe's \emph{activation pattern} $a = \operatorname{Cov}(h)\, w_{\mathrm{raw}}$, and more directly by the class-mean difference $\Delta\mu = \mathbb{E}[h \mid y = 1] - \mathbb{E}[h \mid y = 0]$. We compute all three vectors on each probe's own training states.

Across the 150 probes (25 tasks, three layers, two models), the pattern and the class-mean difference are almost the same vector (median cosine $0.99$ for both models), while the weight vector is far from both (median cosine with the pattern $0.30$ for \llama{} and $0.23$ for \qwen{}). The weight vector is also less stable. Each probe is trained five times, once for each held-out fold, and we take the least similar pair of its five versions: the median cosine of this pair is $0.82$ for \llama{} and $0.84$ for \qwen{} for the weight vector, against $0.99$ for both the pattern and the class-mean difference. The direction along which recovering and failed states differ is thus a stable property of each task and layer, whereas the weight vector varies with the questions the probe is trained on.

\FloatBarrier
\section{Task-Solving Directions}
\label{app:directions}

\subsection{Rotation Fraction, Norm and Layer}
\label{app:directions-robustness}

\subimport{Tables/}{robustness}

Table~\ref{tab:robustness} varies the intervention around the configuration of Section~\ref{sec:direction-method}, at layer~29 of \llama{} and layer~25 of \qwen{}, and Table~\ref{tab:layers} varies the layer.

\paragraph{Rotation fraction.} Accuracy rises with the rotation fraction $t$ and is highest at $t = 0.75$ and $t = 1$. At $t \le 0.5$ the state stays close to its original direction, and the model still answers within a few tokens; past $t = 1$ the state moves beyond the target and accuracy falls slightly.

\paragraph{Direction versus norm.} The rotation keeps the norm of the state and changes only its direction. To separate the two, we also replace the state with the target scaled to $s$ times its own norm, $z^{\ell} \leftarrow s\, d_q(0.5)$, so that $s$ changes only the norm. Since the rotation uses only the direction of its target, the norm $\|c^{+}_{q}\|$ in Eq.~\eqref{eq:direction} matters only for this replacement and for the average that gives $d_T$, and $s = 1$ inserts a state with the norm of the recovering mean, whereas $t = 1$ keeps the norm of the original state. Accuracy depends on the direction far more than on the norm. Rotating further towards the target raises accuracy from 54 to 85 on \llama{} and from 56 to 86 on \qwen{}, whereas scaling the target anywhere from half to four times its norm keeps accuracy between 83 and 87. Accuracy drops only at $s = 0.25$, where the inserted state is much smaller than the state it replaces and too small to have its full effect.

\paragraph{Rotation at every token.} The rotation changes one state, once. Applied at every generated token instead, as steering vectors usually are, the same rotation towards the recovering mean helps far less: at its best it reaches 53\% and 59\%, and at $t = 1$ accuracy falls to 6\% for both models while the generations grow to over a hundred tokens, close to the limit of 128 (Appendix~\ref{sec:implementation}). A single change to the state that generation starts from is therefore enough for the model to compute the answer itself, and changing every later state as well overrides this computation.

\subimport{Tables/}{layers}

\paragraph{Layer.} Table~\ref{tab:layers} rotates the state towards the recovering mean $c^{+}_{q}$ at each layer of the late band. Layer~29 is the best of the three for \llama{}, while for \qwen{} layer~27 gives three points more than layer~25.

\subsection{Per-Task Results}
\label{app:directions-per-task}

Table~\ref{tab:directions-per-task} gives the accuracy on every task under the recovering mean, $d_q(0.5)$, the failed mean and the task-level direction $d_T(0.5)$. The recovering mean and $d_q(0.5)$ raise accuracy above the direct prompt on every task for both models. Over the 25 tasks, $d_q(0.5)$ reaches a higher accuracy than CoT for both models ($p < 10^{-6}$ for \llama{} and $p = 0.004$ for \qwen{}; two-sided Wilcoxon signed-rank). The failed mean lowers accuracy on every task except Multistep Arithmetic for \llama{}, which the direct prompt already fails, and on six tasks of \llama{} it leaves at most 2\% of the questions right. The task-level direction gains more than ten points over the direct prompt only on Multistep Arithmetic and GSM8K, for both models.

\subsection{Example Generations}
\label{app:examples}

Table~\ref{tab:examples} shows the generations of \llama{} for eight questions. On the fifth and sixth, both prompts answer incorrectly and the direction reaches the correct answer with a short computation. On the last two, CoT reasons its way to a wrong answer.

\subsection{Task-Level Directions Across Tasks}
\label{app:task-transfer}

\begin{figure*}[!ht]
  \centering
  \includegraphics[width=\textwidth]{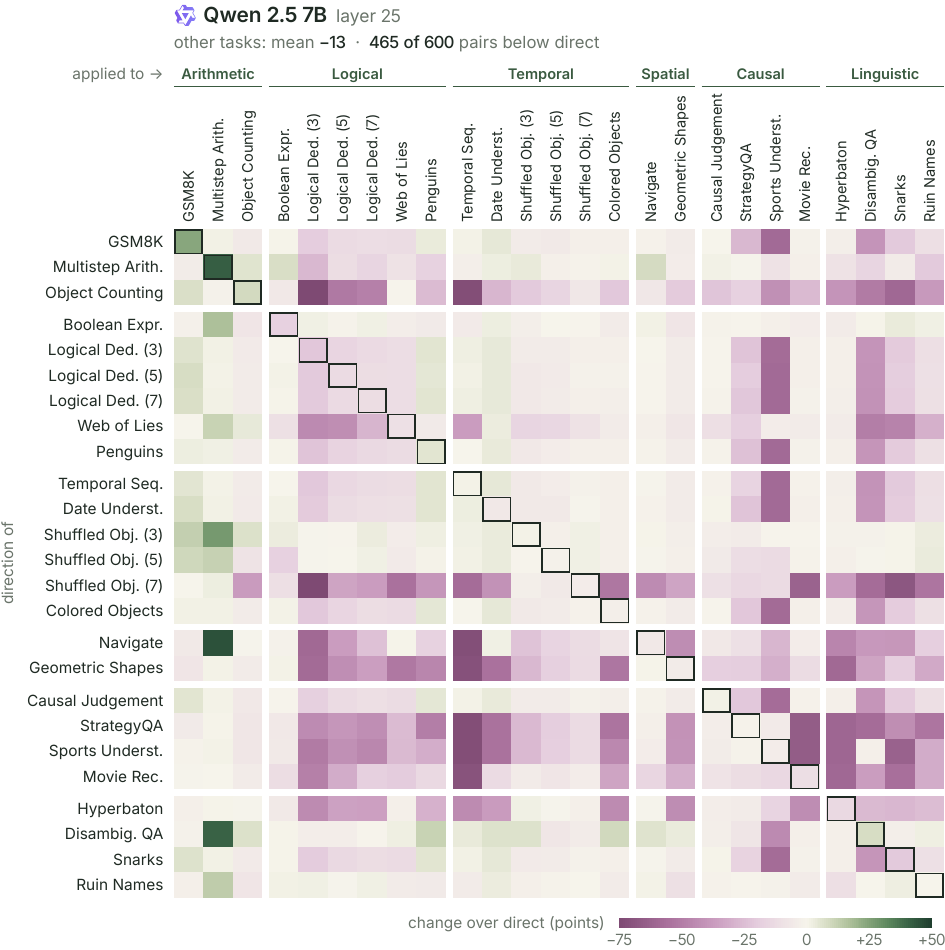}
  \caption{The task-level direction $d_T(0.5)$ of each task (rows) applied to the questions of each task (columns) on \qwen{}, at layer~25 ($t = 1$). Each cell is the change in accuracy over the direct prompt on the column's task, in points. Outlined cells on the diagonal apply a task's direction to its own questions (Figure~\ref{fig:task-level}).}
  \label{fig:task-transfer-qwen}
\end{figure*}

Section~\ref{sec:task-directions} applies each task-level direction to the questions of its own task. Here we apply the task-level direction $d_T(0.5)$ of every task to the questions of every other task on \qwen{} (Figure~\ref{fig:task-transfer-qwen}). Accuracy falls below the direct prompt for 465 of the 600 task pairs, by 18 points on average over these pairs and by as much as 76. The losses concentrate in a few columns: the questions of Sports Understanding, Disambiguation QA and Logical Deduction (3) lose 34, 32 and 30 points on average under the directions of the other tasks.

\subimport{Tables/}{directions-per-task}
\subimport{Tables/}{examples}

\end{document}

%% file: Tables/baselines.tex
\begin{table}[!t]
\centering
\footnotesize
\renewcommand{\arraystretch}{1.06}
\setlength{\tabmodelgap}{0.9em}%
\settowidth{\tabdeltawd}{\textminus00}%
\begin{tabular*}{\columnwidth}{@{\extracolsep{\fill}}>{\hspace{0.55em}}l
    @{\hspace{0.2em}}r@{\hspace{0.45em}}r@{\hspace{0.35em}}w{r}{\tabdeltawd}@{\hspace{0.65em}}r
    @{\hspace{\tabmodelgap}}r@{\hspace{0.45em}}r@{\hspace{0.35em}}w{r}{\tabdeltawd}@{\hspace{0.65em}}r@{}}
\toprule
 & \multicolumn{4}{@{}c@{\hspace{\tabmodelgap}}}{\modelwithlogo{icons/meta-color.pdf}{\llama}}
 & \multicolumn{4}{@{}c@{}}{\modelwithlogo{icons/qwen-color.pdf}{\qwen}} \\
\cmidrule(r{\tabmodelgap}){2-5}\cmidrule{6-9}
\multicolumn{1}{@{}l}{\colhead{Task}} & \colhead{CoT} & \colhead{Dir.} & \colhead{$\Delta$} & \colhead{Tok.} & \colhead{CoT} & \colhead{Dir.} & \colhead{$\Delta$} & \colhead{Tok.} \\
\midrule
\addlinespace[0.25em]
\taskgroup{Arithmetic} \\[-1.2pt]
GSM8K & 85 & 9 & \dpos{76} & 179 & 90 & 20 & \dpos{70} & 248 \\
Multistep Arith. & 66 & 0 & \dpos{65} & 140 & 89 & 2 & \dpos{87} & 260 \\
Object Counting & 80 & 41 & \dpos{38} & 95 & 79 & 49 & \dpos{30} & 117 \\
\groupskip
\taskgroup{Logical} \\[-1.2pt]
Boolean Expr. & 84 & 58 & \dpos{26} & 138 & 95 & 72 & \dpos{23} & 169 \\
Logical Ded.\ (3) & 74 & 51 & \dpos{23} & 128 & 88 & 76 & \dpos{12} & 174 \\
Logical Ded.\ (5) & 51 & 37 & \dpos{14} & 265 & 65 & 54 & \dpos{12} & 333 \\
Logical Ded.\ (7) & 42 & 30 & \dpos{12} & 383 & 52 & 50 & \dpos{2} & 426 \\
Web of Lies & 92 & 51 & \dpos{40} & 120 & 95 & 58 & \dpos{37} & 187 \\
Penguins & 86 & 42 & \dpos{44} & 158 & 97 & 52 & \dpos{45} & 146 \\
\groupskip
\taskgroup{Temporal and state tracking} \\[-1.2pt]
Temporal Seq. & 89 & 78 & \dpos{11} & 284 & 79 & 73 & \dpos{6} & 288 \\
Date Underst. & 72 & 47 & \dpos{24} & 221 & 75 & 57 & \dpos{18} & 223 \\
Shuffled Obj.\ (3) & 63 & 33 & \dpos{30} & 136 & 82 & 27 & \dpos{54} & 214 \\
Shuffled Obj.\ (5) & 58 & 16 & \dpos{42} & 210 & 84 & 19 & \dpos{64} & 321 \\
Shuffled Obj.\ (7) & 45 & 16 & \dpos{29} & 307 & 74 & 13 & \dpos{61} & 467 \\
Colored Objects & 78 & 56 & \dpos{22} & 111 & 87 & 62 & \dpos{26} & 133 \\
\groupskip
\taskgroup{Spatial and geometric} \\[-1.2pt]
Navigate & 79 & 60 & \dpos{20} & 182 & 87 & 68 & \dpos{19} & 214 \\
Geometric Shapes & 41 & 18 & \dpos{22} & 413 & 61 & 44 & \dpos{18} & 308 \\
\groupskip
\taskgroup{Causal, commonsense, and knowledge} \\[-1.2pt]
Causal Judgement & 59 & 53 & \dpos{6} & 160 & 55 & 56 & \dneg{1} & 217 \\
StrategyQA & 75 & 64 & \dpos{11} & 232 & 74 & 69 & \dpos{5} & 262 \\
Sports Underst. & 79 & 74 & \dpos{5} & 143 & 75 & 70 & \dpos{5} & 223 \\
Movie Rec. & 53 & 63 & \dneg{10} & 258 & 60 & 66 & \dneg{6} & 381 \\
\groupskip
\taskgroup{Linguistic and pragmatic} \\[-1.2pt]
Hyperbaton & 78 & 74 & \dpos{4} & 203 & 83 & 60 & \dpos{22} & 197 \\
Disambig.\ QA & 57 & 35 & \dpos{22} & 134 & 66 & 58 & \dpos{9} & 210 \\
Snarks & 71 & 62 & \dpos{9} & 146 & 76 & 74 & \dpos{3} & 201 \\
Ruin Names & 62 & 71 & \dneg{9} & 175 & 51 & 54 & \dneg{3} & 215 \\
\addlinespace[0.25em]
\midrule
\multicolumn{1}{@{}l}{Mean (25 tasks)} & 69 & 46 & \dpos{23} & 197 & 77 & 52 & \dpos{25} & 245 \\
\bottomrule
\end{tabular*}
\caption{CoT and direct (Dir.\@) accuracy (\%) on the 25 tasks, with the CoT gain $\Delta$ (computed before rounding) and the mean number of tokens generated per question under CoT (direct prompting generates about one).}
\label{tab:baselines}
\end{table}

%% file: Tables/prompts.tex
\begin{table}[t]
\centering
\footnotesize
\renewcommand{\arraystretch}{1.12}
\begin{tabular*}{\columnwidth}{@{\extracolsep{\fill}}l@{\hspace{0.6em}}p{0.68\columnwidth}@{}}
\toprule
\multicolumn{2}{@{}l}{\colhead{Template, shared across tasks}} \\
\midrule
User turn & \texttt{Question: \{question\}} \texttt{Answer:} \\
System, CoT & \texttt{you are a step-by-step reasoner, finish the reasoning with Final Answer: \#\#\#\# $\langle$ans$\rangle$} \\
System, direct & \texttt{Answer only with $\langle$ans$\rangle$. Do not show steps.} \\
\addlinespace[0.4em]
\multicolumn{2}{@{}l}{\colhead{Answer slot $\langle$ans$\rangle$ (CoT; direct)}} \\
\midrule
True/False & \texttt{[True/False]}; \texttt{True or False} \\
Yes/No & \texttt{[Yes/No]}; \texttt{Yes or No} \\
Multiple choice & \texttt{[single letter, e.g.\ A]}; \texttt{a single letter (e.g.\ A, B, C)} \\
GSM8K & \texttt{[number]}; \texttt{the final numeric result} \\
Numeric (BBH) & \texttt{[number]}; \texttt{a number} \\
\bottomrule
\end{tabular*}
\caption{Prompt templates. Every task uses the same user turn, and the two system instructions differ only in whether they ask for steps and in the answer slot, which depends on the answer format. True/False covers StrategyQA and the True/False tasks of BIG-Bench Hard. Under CoT, \texttt{\#\#\#\#} marks the final answer.}
\label{tab:prompts}
\end{table}

%% file: Tables/dataset.tex
\begin{table*}[!t]
\centering
\small
\renewcommand{\arraystretch}{1.1}
\begin{minipage}{0.8\textwidth}
\setlength{\tabcolsep}{0pt}%
\begin{tabular*}{\linewidth}{@{\extracolsep{\fill}}>{\hspace{0.6em}}l r@{\hspace{1.2em}}r r r@{\hspace{1.8em}}r r r@{}}
\toprule
 & & \multicolumn{3}{c}{\modelwithlogo{icons/meta-color.pdf}{\llama}}
 & \multicolumn{3}{c@{}}{\modelwithlogo{icons/qwen-color.pdf}{\qwen}} \\
\cmidrule(lr){3-5}\cmidrule(l){6-8}
\multicolumn{1}{@{}l}{\colhead{Task}} & \colhead{Questions} & \colhead{States} & \colhead{Rec.\ (\%)} & \colhead{Cov.\ (\%)} & \colhead{States} & \colhead{Rec.\ (\%)} & \colhead{Cov.\ (\%)} \\
\midrule
\addlinespace[0.25em]
\taskgroupn{8}{Arithmetic} \\[-1.2pt]
GSM8K & 1{,}319 & 172 & 26 & 97 & 246 & 30 & 98 \\
Multistep Arith. & 250 & 142 & 11 & 99 & 262 & 10 & 100 \\
Object Counting & 250 & 94 & 23 & 94 & 119 & 34 & 98 \\
\groupskip
\taskgroupn{8}{Logical} \\[-1.2pt]
Boolean Expr. & 250 & 141 & 47 & 100 & 174 & 56 & 100 \\
Logical Ded.\ (3) & 250 & 130 & 17 & 98 & 174 & 29 & 99 \\
Logical Ded.\ (5) & 250 & 251 & 12 & 96 & 322 & 20 & 94 \\
Logical Ded.\ (7) & 250 & 293 & 11 & 84 & 364 & 19 & 94 \\
Web of Lies & 250 & 122 & 13 & 99 & 187 & 31 & 100 \\
Penguins & 146 & 146 & 14 & 100 & 147 & 23 & 99 \\
\groupskip
\taskgroupn{8}{Temporal and state tracking} \\[-1.2pt]
Temporal Seq. & 250 & 277 & 16 & 100 & 287 & 23 & 100 \\
Date Underst. & 250 & 173 & 11 & 94 & 218 & 22 & 98 \\
Shuffled Obj.\ (3) & 250 & 138 & 16 & 99 & 215 & 12 & 97 \\
Shuffled Obj.\ (5) & 250 & 203 & 10 & 95 & 312 & 8 & 98 \\
Shuffled Obj.\ (7) & 250 & 274 & 8 & 89 & 379 & 7 & 80 \\
Colored Objects & 250 & 114 & 17 & 94 & 133 & 32 & 99 \\
\groupskip
\taskgroupn{8}{Spatial and geometric} \\[-1.2pt]
Navigate & 250 & 180 & 22 & 100 & 216 & 53 & 100 \\
Geometric Shapes & 250 & 341 & 4 & 84 & 293 & 14 & 88 \\
\groupskip
\taskgroupn{8}{Causal, commonsense, and knowledge} \\[-1.2pt]
Causal Judgement & 187 & 166 & 20 & 91 & 217 & 45 & 95 \\
StrategyQA & 2{,}290 & 227 & 19 & 97 & 263 & 51 & 100 \\
Sports Underst. & 250 & 143 & 15 & 98 & 224 & 39 & 99 \\
Movie Rec. & 250 & 249 & 22 & 95 & 369 & 29 & 99 \\
\groupskip
\taskgroupn{8}{Linguistic and pragmatic} \\[-1.2pt]
Hyperbaton & 250 & 200 & 31 & 100 & 198 & 28 & 98 \\
Disambig.\ QA & 250 & 136 & 10 & 94 & 211 & 27 & 100 \\
Snarks & 178 & 145 & 20 & 100 & 201 & 29 & 100 \\
Ruin Names & 250 & 172 & 22 & 99 & 216 & 30 & 99 \\
\addlinespace[0.25em]
\midrule
\multicolumn{1}{@{}l}{All 25 tasks} & 9{,}120 & 185 & 17 & 96 & 238 & 28 & 97 \\
\bottomrule
\end{tabular*}
\end{minipage}
\caption{Patch labels per task, at layer~29 of \llama{} and layer~25 of \qwen{}. Questions is the number of questions; States is the mean number of labelled CoT states per question, one per token of the chain, which is capped at 400 tokens (Appendix~\ref{sec:implementation}); Rec.\ is the percentage of these states that recover the answer when patched; Cov.\ is the percentage of questions with at least one recovering and one failed state, for which the directions of Section~\ref{sec:directions} are defined. The last row gives the total number of questions and, for the other columns, macro averages over tasks.}
\label{tab:dataset}
\end{table*}

%% file: Tables/appendix-probe-selectivity-layers.tex
\begin{table}[!htb]
  \centering
  \small
  \setlength{\tabcolsep}{3pt}
  \begin{tabular}{@{}lccccc@{}}
    \toprule
    & & \multicolumn{2}{c}{Control} & \multicolumn{2}{c}{Selectivity} \\
    \cmidrule(lr){3-4} \cmidrule(lr){5-6}
    $\ell$ & Probe & Orig. & W.-pos. & Orig. & W.-pos. \\
    \midrule
    \multicolumn{6}{@{}l}{\modelwithlogo{icons/meta-color.pdf}{\llama{}}} \\
    27 & 0.819 & 0.625 & 0.551 & 0.194 & 0.267 \\
    29 & 0.836 & 0.615 & 0.552 & 0.221 & 0.284 \\
    31 & 0.838 & 0.620 & 0.552 & 0.218 & 0.286 \\
    \midrule
    \multicolumn{6}{@{}l}{\modelwithlogo{icons/qwen-color.pdf}{\qwen{}}} \\
    23 & 0.795 & 0.601 & 0.534 & 0.194 & 0.261 \\
    25 & 0.796 & 0.599 & 0.532 & 0.197 & 0.263 \\
    27 & 0.794 & 0.602 & 0.529 & 0.192 & 0.264 \\
    \bottomrule
  \end{tabular}
\caption{Probe performance on the original and control tasks at each layer of the late band, summarised over the 25 tasks. Probe: mean out-of-fold AUROC, computed over all questions of each task. Control: mean AUROC of the origin (Orig.) and within-position (W.-pos.) control probes (Appendix~\ref{app:probe-controls}). Selectivity: probe AUROC minus control AUROC.}
  \label{tab:probe-layers}
\end{table}

%% file: Tables/robustness.tex
\begin{table}[!ht]
\centering
\footnotesize
\renewcommand{\arraystretch}{1.06}
\setlength{\tabmodelgap}{1.2em}%
\begin{tabular*}{\columnwidth}{@{\extracolsep{\fill}}>{\hspace{0.45em}}l
    @{\hspace{0.6em}}r@{\hspace{0.6em}}r@{\hspace{\tabmodelgap}}r@{\hspace{0.6em}}r@{}}
\toprule
 & \multicolumn{2}{@{}c@{\hspace{\tabmodelgap}}}{\modelwithlogo{icons/meta-color.pdf}{\llama}}
 & \multicolumn{2}{@{}c@{}}{\modelwithlogo{icons/qwen-color.pdf}{\qwen}} \\
\cmidrule(r{\tabmodelgap}){2-3}\cmidrule{4-5}
 & \colhead{Acc.} & \colhead{Tok.} & \colhead{Acc.} & \colhead{Tok.} \\
\midrule
\addlinespace[0.25em]
\taskgroupn{5}{Rotation fraction $t$, towards $d_{q}(0.5)$} \\[-1.2pt]
$t = 0.25$ & 54 & 1 & 56 & 1 \\
$t = 0.5$ & 71 & 5 & 67 & 3 \\
$t = 0.75$ & 85 & 16 & 86 & 15 \\
$t = 1$ & 85 & 19 & 86 & 19 \\
$t = 1.25$ & 82 & 19 & 84 & 20 \\
\groupskip
\taskgroupn{5}{Target inserted at $s$ times its norm, $d_{q}(0.5)$} \\[-1.2pt]
$s = 0.25$ & 70 & 13 & 60 & 15 \\
$s = 0.5$ & 83 & 18 & 83 & 19 \\
$s = 1$ & 85 & 19 & 87 & 19 \\
$s = 2$ & 84 & 18 & 87 & 19 \\
$s = 4$ & 84 & 18 & 86 & 19 \\
\groupskip
\taskgroupn{5}{At every generated token, towards $c^{+}_{q}$} \\[-1.2pt]
$t = 0.25$ & 50 & 1 & 54 & 1 \\
$t = 0.5$ & 53 & 64 & 59 & 9 \\
$t = 1$ & 6 & 111 & 6 & 109 \\
\addlinespace[0.25em]
\bottomrule
\end{tabular*}
\caption{Variations of the intervention, macro-averaged over the 25 tasks: accuracy in percent and mean tokens generated per question. Unless the block says otherwise, the last state of the direct prompt is rotated once, at layer~29 of \llama{} and layer~25 of \qwen{}. The main text uses $t = 1$ towards $d_{q}(0.5)$, which gives 85\% for \llama{} and 86\% for \qwen{}.}
\label{tab:robustness}
\end{table}

%% file: Tables/layers.tex
\begin{table}[!ht]
\centering
\footnotesize
\renewcommand{\arraystretch}{1.06}
\setlength{\tabmodelgap}{1.2em}%
\begin{tabular*}{\columnwidth}{@{\extracolsep{\fill}}c
    @{\hspace{0.6em}}r@{\hspace{0.6em}}r@{\hspace{\tabmodelgap}}c@{\hspace{0.6em}}r@{\hspace{0.6em}}r@{}}
\toprule
\multicolumn{3}{@{}c@{\hspace{\tabmodelgap}}}{\modelwithlogo{icons/meta-color.pdf}{\llama}}
 & \multicolumn{3}{@{}c@{}}{\modelwithlogo{icons/qwen-color.pdf}{\qwen}} \\
\cmidrule(r{\tabmodelgap}){1-3}\cmidrule{4-6}
\colhead{Layer} & \colhead{Acc.} & \colhead{Tok.} & \colhead{Layer} & \colhead{Acc.} & \colhead{Tok.} \\
\midrule
27 & 75 & 16 & 23 & 76 & 18 \\
29 & 80 & 17 & 25 & 82 & 18 \\
31 & 76 & 15 & 27 & 85 & 18 \\
\bottomrule
\end{tabular*}
\caption{Rotating the last state of the direct prompt once, at $t = 1$, towards $c^{+}_{q}$ at each layer of the late band; columns as in Table~\ref{tab:robustness}. The main text uses layer~29 of \llama{} and layer~25 of \qwen{}.}
\label{tab:layers}
\end{table}

%% file: Tables/directions-per-task.tex
\begin{table*}[p]
\centering
\footnotesize
\renewcommand{\arraystretch}{1.0}
\setlength{\tabmodelgap}{1.6em}%
\begin{tabular*}{\textwidth}{@{\extracolsep{\fill}}>{\hspace{0.45em}}l
    @{\hspace{0.6em}}r@{\hspace{0.6em}}r@{\hspace{0.6em}}r@{\hspace{0.6em}}r@{\hspace{0.6em}}r@{\hspace{0.6em}}r@{\hspace{0.9em}}r
    @{\hspace{\tabmodelgap}}r@{\hspace{0.6em}}r@{\hspace{0.6em}}r@{\hspace{0.6em}}r@{\hspace{0.6em}}r@{\hspace{0.6em}}r@{\hspace{0.9em}}r@{}}
\toprule
 & \multicolumn{7}{@{}c@{\hspace{\tabmodelgap}}}{\modelwithlogo{icons/meta-color.pdf}{\llama}}
 & \multicolumn{7}{@{}c@{}}{\modelwithlogo{icons/qwen-color.pdf}{\qwen}} \\
\cmidrule(r{\tabmodelgap}){2-8}\cmidrule{9-15}
\multicolumn{1}{@{}l}{\colhead{Task}} & \colhead{Dir.} & \colhead{CoT} & \colhead{$c^{+}_{q}$} & \colhead{$d_{q}(0.5)$} & \colhead{$c^{-}_{q}$} & \colhead{$d_{T}(0.5)$} & \colhead{Tok.} & \colhead{Dir.} & \colhead{CoT} & \colhead{$c^{+}_{q}$} & \colhead{$d_{q}(0.5)$} & \colhead{$c^{-}_{q}$} & \colhead{$d_{T}(0.5)$} & \colhead{Tok.} \\
\midrule
\addlinespace[0.25em]
\taskgroupn{15}{Arithmetic} \\[-1.2pt]
GSM8K & 9 & 85 & 73 & \textbf{86} & 6 & 32 & 39 & 20 & \textbf{90} & 74 & 83 & 9 & 44 & 52 \\
Multistep Arith. & 0 & 66 & 79 & \textbf{82} & 1 & 48 & 55 & 2 & \textbf{89} & 67 & 74 & 0 & 44 & 66 \\
Object Counting & 41 & 80 & 86 & \textbf{89} & 7 & 40 & 11 & 49 & 79 & 78 & \textbf{87} & 20 & 58 & 12 \\
\groupskip
\taskgroupn{15}{Logical} \\[-1.2pt]
Boolean Expr. & 58 & 84 & 96 & \textbf{100} & 7 & 54 & 3 & 72 & 95 & 98 & \textbf{100} & 13 & 54 & 2 \\
Logical Ded.\ (3) & 51 & 74 & 81 & \textbf{87} & 6 & 45 & 14 & 76 & 88 & 85 & \textbf{91} & 11 & 55 & 9 \\
Logical Ded.\ (5) & 37 & 51 & 83 & \textbf{86} & 7 & 41 & 23 & 54 & 65 & 80 & \textbf{85} & 16 & 41 & 17 \\
Logical Ded.\ (7) & 30 & 42 & 67 & \textbf{70} & 3 & 39 & 22 & 50 & 52 & 70 & \textbf{78} & 16 & 38 & 21 \\
Web of Lies & 51 & 92 & 92 & \textbf{92} & 8 & 48 & 6 & 58 & \textbf{95} & 83 & 87 & 11 & 48 & 40 \\
Penguins & 42 & \textbf{86} & 88 & 85 & 10 & 29 & 16 & 52 & \textbf{97} & 88 & 92 & 13 & 58 & 17 \\
\groupskip
\taskgroupn{15}{Temporal and state tracking} \\[-1.2pt]
Temporal Seq. & 78 & 89 & 94 & \textbf{96} & 48 & 82 & 10 & 73 & 79 & 91 & \textbf{94} & 64 & 74 & 17 \\
Date Underst. & 47 & 72 & 77 & \textbf{83} & 9 & 50 & 18 & 57 & 75 & 79 & \textbf{85} & 7 & 51 & 21 \\
Shuffled Obj.\ (3) & 33 & 63 & 82 & \textbf{86} & 5 & 25 & 37 & 27 & 82 & 83 & \textbf{85} & 8 & 28 & 15 \\
Shuffled Obj.\ (5) & 16 & 58 & 73 & \textbf{72} & 8 & 19 & 45 & 19 & \textbf{84} & 78 & 77 & 3 & 17 & 24 \\
Shuffled Obj.\ (7) & 16 & 45 & 56 & \textbf{56} & 8 & 18 & 51 & 13 & \textbf{74} & 62 & 62 & 1 & 9 & 24 \\
Colored Objects & 56 & 78 & 81 & \textbf{84} & 6 & 55 & 13 & 62 & 87 & 88 & \textbf{92} & 16 & 58 & 16 \\
\groupskip
\taskgroupn{15}{Spatial and geometric} \\[-1.2pt]
Navigate & 60 & 79 & 87 & \textbf{93} & 51 & 57 & 12 & 68 & 87 & 88 & \textbf{91} & 34 & 62 & 26 \\
Geometric Shapes & 18 & 41 & 65 & \textbf{61} & 1 & 22 & 18 & 44 & 61 & 72 & \textbf{68} & 6 & 39 & 13 \\
\groupskip
\taskgroupn{15}{Causal, commonsense, and knowledge} \\[-1.2pt]
Causal Judgement & 53 & 59 & 76 & \textbf{83} & 11 & 55 & 10 & 56 & 55 & 81 & \textbf{86} & 28 & 57 & 20 \\
StrategyQA & 64 & 75 & 74 & \textbf{87} & 5 & 45 & 9 & 69 & 74 & 86 & \textbf{92} & 29 & 68 & 13 \\
Sports Underst. & 74 & 79 & 80 & \textbf{86} & 10 & 65 & 6 & 70 & 75 & 83 & \textbf{92} & 14 & 66 & 9 \\
Movie Rec. & 63 & 53 & 77 & \textbf{86} & 0 & 56 & 13 & 66 & 60 & 88 & \textbf{94} & 4 & 54 & 7 \\
\groupskip
\taskgroupn{15}{Linguistic and pragmatic} \\[-1.2pt]
Hyperbaton & 74 & 78 & 94 & \textbf{96} & 0 & 74 & 9 & 60 & 83 & 96 & \textbf{96} & 17 & 47 & 4 \\
Disambig.\ QA & 35 & 57 & 75 & \textbf{84} & 1 & 38 & 11 & 58 & 66 & 82 & \textbf{90} & 14 & 66 & 11 \\
Snarks & 62 & 71 & 90 & \textbf{95} & 3 & 72 & 5 & 74 & 76 & 82 & \textbf{93} & 15 & 53 & 9 \\
Ruin Names & 71 & 62 & 74 & \textbf{93} & 1 & 42 & 8 & 54 & 51 & 82 & \textbf{89} & 21 & 54 & 12 \\
\addlinespace[0.25em]
\midrule
\multicolumn{1}{@{}l}{Mean (25 tasks)} & 46 & 69 & 80 & 85 & 9 & 46 & 19 & 52 & 77 & 82 & 86 & 16 & 50 & 19 \\
\bottomrule
\end{tabular*}
\caption{Accuracy in percent per task after rotating the state of the direct prompt ($t = 1$; layer~29 of \llama{}, layer~25 of \qwen{}) towards the question's recovering mean $c^{+}_{q}$, contrastive direction $d_{q}(0.5)$ and failed mean $c^{-}_{q}$, and towards the task-level direction $d_{T}(0.5)$ of its task, against the direct (Dir.) and CoT prompts. Tok.\ is the mean number of tokens generated under $d_{q}(0.5)$. Questions without a contrastive direction count as incorrect. The higher of CoT and $d_{q}(0.5)$ is in bold, compared before rounding.}
\label{tab:directions-per-task}
\end{table*}

%% file: Tables/examples.tex
\begin{table*}[!t]
\centering
\footnotesize

\definecolor{exGroup}{HTML}{6D6C6C}  
\definecolor{exRule}{HTML}{C4C4C2}   
\definecolor{exDir}{HTML}{8A8F86}    
\definecolor{exVec}{HTML}{344E41}    
\definecolor{exCot}{HTML}{99582A}    
\definecolor{exOk}{HTML}{2B9C8D}     
\definecolor{exNo}{HTML}{D1495B}     
\newcommand{\exsans}{\fontfamily{phv}\fontsize{8}{9.5}\selectfont}
\newcommand{\exlab}[2]{\leavevmode{\exsans\color{#1}#2}}
\newcommand{\extag}[1]{\unskip\nobreak\hfil\penalty50\hskip1.5em\hbox{}\nobreak\hfill#1\parfillskip=0pt\finalhyphendemerits=0\relax}
\newcommand{\exok}{\extag{\exsans\color{exOk}correct}}
\newcommand{\exno}{\extag{\exsans\color{exNo}wrong}}
\newcommand{\exrow}[2]{#1 & #2 \\[0.15em]}
\newcommand{\exm}{\textminus}
\newcommand{\ext}{\textasteriskcentered}
\newcommand{\exhead}[1]{\multicolumn{2}{@{}l@{}}{\makebox[\linewidth][l]{\exsans\color{exGroup}#1\hspace{0.5em}{\color{exRule}\leaders\hrule height 0.62ex depth -0.56ex\hfill}}}}
\newenvironment{exbox}[1]{%
  \begin{minipage}[t]{0.483\textwidth}
  \renewcommand{\arraystretch}{1.08}%
  \begin{tabular*}{\linewidth}[t]{@{}p{3.75em}@{\hspace{0.45em}}>{\raggedright\arraybackslash}p{\dimexpr\linewidth-4.2em\relax}@{}}
  \exhead{#1} \\[0.2em]}{%
  \end{tabular*}
  \end{minipage}}
\noindent\rule{\textwidth}{0.08em}\\[0.45em]
\begin{exbox}{GSM8K}
\exrow{\exlab{exGroup}{Question}}{Ali had \$21. Leila gave him half of her \$100. How much does Ali have now?}
\exrow{\exlab{exDir}{Direct}}{121 \exno}
\exrow{\exlab{exVec}{$d_{q}(0.5)$}}{add \$21 and \$50. \$21 + \$50 = \$71 \exok}
\exrow{\exlab{exCot}{CoT}}{Let's break it down step by step: 1. Ali initially had \$21. 2. Leila gave him half of her \$100, which is \$100 / 2 = \$50. \ldots{} \mbox{Final Answer: \$71} \exok}
\end{exbox}\hfill
\begin{exbox}{Web of Lies}
\exrow{\exlab{exGroup}{Question}}{Tamika lies. Raymond says Tamika tells the truth. Willian says Raymond tells the truth. Shaunda says Willian lies. Elanor says Shaunda tells the truth. Does Elanor tell the truth?}
\exrow{\exlab{exDir}{Direct}}{No \exno}
\exrow{\exlab{exVec}{$d_{q}(0.5)$}}{Yes \exok}
\exrow{\exlab{exCot}{CoT}}{Let's break it down step by step: 1. Raymond says Tamika tells the truth, but we know Tamika lies, so Raymond must be lying. \ldots{} \mbox{Final Answer: Yes} \exok}
\end{exbox}\\[0.8em]
\begin{exbox}{Temporal Sequences}
\exrow{\exlab{exGroup}{Question}}{Today, Susan went to the coffee shop. Between what times could they have gone? We know that: Susan woke up at 7am. Linda saw Susan driving to the water park from 7am to 11am \ldots}
\exrow{\exlab{exDir}{Direct}}{C \exno}
\exrow{\exlab{exVec}{$d_{q}(0.5)$}}{the correct answer is A. \exok}
\exrow{\exlab{exCot}{CoT}}{Let's analyze the information step by step: 1. Susan woke up at 7am, so she was definitely available to go to the coffee shop between 7am and 9pm. \ldots{} \mbox{Final Answer: A} \exok}
\end{exbox}\hfill
\begin{exbox}{Snarks}
\exrow{\exlab{exGroup}{Question}}{Which statement is sarcastic? Options: (A) Hey just be happy then you won't be depressed anymore (B) Hey just be happy that you won't be depressed anymore}
\exrow{\exlab{exDir}{Direct}}{B \exno}
\exrow{\exlab{exVec}{$d_{q}(0.5)$}}{of (A) \exok}
\exrow{\exlab{exCot}{CoT}}{To determine which statement is sarcastic, let's analyze the options: (A) ``Hey just be happy then you won't be depressed anymore'' -- This statement implies that happiness is a simple solution to depression \ldots{} \mbox{Final Answer: A} \exok}
\end{exbox}\\[0.8em]
\begin{exbox}{Multistep Arithmetic}
\exrow{\exlab{exGroup}{Question}}{((\exm4 + \exm6 + \exm6 \exm{} \exm9) + (\exm2 \ext{} \exm5 \ext{} 5 \ext{} 7)) =}
\exrow{\exlab{exDir}{Direct}}{\exm2 \exno}
\exrow{\exlab{exVec}{$d_{q}(0.5)$}}{parentheses first: (\exm4 + \exm6 + \exm6 \exm{} \exm9) = \exm7\newline then multiply: (\exm2 \ext{} \exm5 \ext{} 5 \ext{} 7) = 350\newline then add: \exm7 + 350 = 343 \exok}
\exrow{\exlab{exCot}{CoT}}{To evaluate the expression, we need to follow the order of operations (PEMDAS): \ldots{} \mbox{(\exm4 + \exm6 + \exm6 \exm{} \exm9) = \exm7} \ldots{} \mbox{(\exm2 \ext{} \exm5 \ext{} 5 \ext{} 7) = \exm2 \ext{} 125 = \exm250} \ldots{} \mbox{\strut\exm7 + (\exm250) = \exm257} \mbox{Final Answer: \exm257} \exno}
\end{exbox}\hfill
\begin{exbox}{Object Counting}
\exrow{\exlab{exGroup}{Question}}{I have a trumpet, four trombones, an accordion, a clarinet, a violin, and a drum. How many musical instruments do I have?}
\exrow{\exlab{exDir}{Direct}}{8 \exno}
\exrow{\exlab{exVec}{$d_{q}(0.5)$}}{Clarinet, Drum, Accordion, Trumpet, Violin, Trombone, Trombone, Trombone, \mbox{Trombone = 9} \exok}
\exrow{\exlab{exCot}{CoT}}{Let's count the instruments: 1. Trumpet 2. Four trombones (so that's 4) 3. Accordion 4. Clarinet 5. Violin 6. Drum There are 6 instruments in total. \mbox{Final Answer: 6} \exno}
\end{exbox}\\[0.8em]
\begin{exbox}{Ruin Names}
\exrow{\exlab{exGroup}{Question}}{Which of the following is a humorous edit of this artist or movie name: `counting crows'? Options: (A) mounting crows (B) coufting crows (C) counting crowj (D) counting brows}
\exrow{\exlab{exDir}{Direct}}{D \exok}
\exrow{\exlab{exVec}{$d_{q}(0.5)$}}{(D) \exok}
\exrow{\exlab{exCot}{CoT}}{To determine the humorous edit, let's analyze each option: \ldots{} (D) counting brows -- This option is a play on words, using ``brows'' instead of ``crows'', but it's not as clever as option (A). \mbox{Final Answer: A} \exno}
\end{exbox}\hfill
\begin{exbox}{Movie Recommendation}
\exrow{\exlab{exGroup}{Question}}{Find a movie similar to Rocky, Star Wars Episode IV - A New Hope, Toy Story, The Terminator: Options: (A) Dracula Dead and Loving It (B) Independence Day (C) The Extraordinary Adventures of Ad\`ele Blanc-Sec (D) The American President}
\exrow{\exlab{exDir}{Direct}}{B \exok}
\exrow{\exlab{exVec}{$d_{q}(0.5)$}}{these movies are all action/adventure films, so I'll choose: (B) \exok}
\exrow{\exlab{exCot}{CoT}}{To find a movie similar to Rocky, Star Wars \ldots{} we need to identify a common theme or genre among these films. \ldots{} \mbox{Final Answer: C} \exno}
\end{exbox}\\[0.5em]
\noindent\rule{\textwidth}{0.08em}
\caption{Generations of \llama{} under the direct prompt, with its last state pointed towards $d_{q}(0.5)$ at layer~29 ($t = 1$), and under CoT. In the first four examples the direct prompt is wrong, and the direction reaches the answer of CoT in a few tokens. In the next two, the CoT and direct prompts are both wrong, and the direction reaches the correct answer with a short computation. Over all 25 tasks, both prompts are wrong on 1,867 of the 9,120 questions for \llama{} and on 1,441 for \qwen{}, and the direction reaches the correct answer on 1,192 (64\%) and 1,014 (70\%) of them. The last two come from tasks where CoT falls below the direct prompt: the direct prompt is already correct, the steps of CoT sway the model away from this answer, and the direction keeps it. Questions and CoT outputs are truncated (\ldots) for space.}
\label{tab:examples}
\end{table*}

%% file: custom.bib
@inproceedings{hewitt-liang-2019-designing,
    title = "Designing and Interpreting Probes with Control Tasks",
    author = "Hewitt, John  and
      Liang, Percy",
    editor = "Inui, Kentaro  and
      Jiang, Jing  and
      Ng, Vincent  and
      Wan, Xiaojun",
    booktitle = "Proceedings of the 2019 Conference on Empirical Methods in Natural Language Processing and the 9th International Joint Conference on Natural Language Processing ({EMNLP}-{IJCNLP})",
    month = nov,
    year = "2019",
    address = "Hong Kong, China",
    publisher = "Association for Computational Linguistics",
    url = "https://aclanthology.org/D19-1275/",
    doi = "10.18653/v1/D19-1275",
    pages = "2733--2743"
}

@inproceedings{rimsky-etal-2024-steering,
    title = "Steering {L}lama 2 via Contrastive Activation Addition",
    author = "Rimsky, Nina  and
      Gabrieli, Nick  and
      Schulz, Julian  and
      Tong, Meg  and
      Hubinger, Evan  and
      Turner, Alexander",
    editor = "Ku, Lun-Wei  and
      Martins, Andre  and
      Srikumar, Vivek",
    booktitle = "Proceedings of the 62nd Annual Meeting of the {A}ssociation for {C}omputational {L}inguistics (Volume 1: Long Papers)",
    month = aug,
    year = "2024",
    address = "Bangkok, Thailand",
    publisher = "Association for Computational Linguistics",
    url = "https://aclanthology.org/2024.acl-long.828/",
    doi = "10.18653/v1/2024.acl-long.828",
    pages = "15504--15522"
}

@inproceedings{afzal2025knowing,
    title = "Knowing Before Saying: {LLM} Representations Encode Information About Chain-of-Thought Success Before Completion",
    author = "Afzal, Anum  and
      Matthes, Florian  and
      Chechik, Gal  and
      Ziser, Yftah",
    editor = "Che, Wanxiang  and
      Nabende, Joyce  and
      Shutova, Ekaterina  and
      Pilehvar, Mohammad Taher",
    booktitle = "Findings of the Association for Computational Linguistics: ACL 2025",
    month = jul,
    year = "2025",
    address = "Vienna, Austria",
    publisher = "Association for Computational Linguistics",
    url = "https://aclanthology.org/2025.findings-acl.662/",
    doi = "10.18653/v1/2025.findings-acl.662",
    pages = "12791--12806",
    ISBN = "979-8-89176-256-5"
}

@article{belinkov2022probing,
    title = "Probing Classifiers: Promises, Shortcomings, and Advances",
    author = "Belinkov, Yonatan",
    journal = "Computational Linguistics",
    volume = "48",
    number = "1",
    month = mar,
    year = "2022",
    address = "Cambridge, MA",
    publisher = "MIT Press",
    url = "https://aclanthology.org/2022.cl-1.7/",
    doi = "10.1162/coli_a_00422",
    pages = "207--219"
}

@inproceedings{dang2026selective,
    title = "Selective Steering: Norm-Preserving Control Through Discriminative Layer Selection",
    author = "Dang, Quy-Anh  and
      Ngo, Chris",
    editor = "Liakata, Maria  and
      Moreira, Viviane P.  and
      Zhang, Jiajun  and
      Jurgens, David",
    booktitle = "Findings of the {A}ssociation for {C}omputational {L}inguistics: {ACL} 2026",
    month = jul,
    year = "2026",
    address = "San Diego, California, United States",
    publisher = "Association for Computational Linguistics",
    url = "https://aclanthology.org/2026.findings-acl.529/",
    doi = "10.18653/v1/2026.findings-acl.529",
    pages = "10887--10910",
    ISBN = "979-8-89176-395-1"
}

@article{haufe2014interpretation,
    title = {On the interpretation of weight vectors of linear models in multivariate neuroimaging},
    author = {Haufe, Stefan and Meinecke, Frank and G{\"o}rgen, Kai and D{\"a}hne, Sven and Haynes, John-Dylan and Blankertz, Benjamin and Bie{\ss}mann, Felix},
    journal = {NeuroImage},
    volume = {87},
    pages = {96--110},
    month = feb,
    year = {2014},
    publisher = {Elsevier},
    issn = {1053-8119},
    doi = {10.1016/j.neuroimage.2013.10.067}
}

@article{liu1989limited,
    title = {On the limited memory {BFGS} method for large scale optimization},
    author = {Liu, Dong C. and Nocedal, Jorge},
    journal = {Mathematical Programming},
    volume = {45},
    number = {1--3},
    pages = {503--528},
    month = aug,
    year = {1989},
    publisher = {Springer},
    doi = {10.1007/BF01589116}
}

@article{pedregosa2011scikit,
  author  = {Fabian Pedregosa and Ga{{\"e}}l Varoquaux and Alexandre Gramfort and Vincent Michel and Bertrand Thirion and Olivier Grisel and Mathieu Blondel and Peter Prettenhofer and Ron Weiss and Vincent Dubourg and Jake Vanderplas and Alexandre Passos and David Cournapeau and Matthieu Brucher and Matthieu Perrot and {{\'E}}douard Duchesnay},
  title   = {Scikit-learn: Machine Learning in {P}ython},
  journal = {Journal of Machine Learning Research},
  year    = {2011},
  volume  = {12},
  number  = {85},
  pages   = {2825--2830},
  url     = {http://jmlr.org/papers/v12/pedregosa11a.html}
}

@article{shoemake1985animating,
    title = {Animating rotation with quaternion curves},
    author = {Shoemake, Ken},
    journal = {SIGGRAPH Computer Graphics},
    volume = {19},
    number = {3},
    pages = {245--254},
    month = jul,
    year = {1985},
    publisher = {Association for Computing Machinery},
    issn = {0097-8930},
    doi = {10.1145/325165.325242}
}

@article{strategyqa,
    title = "Did {A}ristotle Use a Laptop? {A} Question Answering Benchmark with Implicit Reasoning Strategies",
    author = "Geva, Mor  and
      Khashabi, Daniel  and
      Segal, Elad  and
      Khot, Tushar  and
      Roth, Dan  and
      Berant, Jonathan",
    editor = "Roark, Brian  and
      Nenkova, Ani",
    journal = "Transactions of the Association for Computational Linguistics",
    volume = "9",
    year = "2021",
    address = "Cambridge, MA",
    publisher = "MIT Press",
    url = "https://aclanthology.org/2021.tacl-1.21/",
    doi = "10.1162/tacl_a_00370",
    pages = "346--361"
}

@inproceedings{kazemi-etal-2025-big,
    title = "{BIG}-{B}ench {E}xtra {H}ard",
    author = "Kazemi, Mehran  and
      Fatemi, Bahare  and
      Bansal, Hritik  and
      Palowitch, John  and
      Anastasiou, Chrysovalantis  and
      Mehta, Sanket Vaibhav  and
      Jain, Lalit K  and
      Aglietti, Virginia  and
      Jindal, Disha  and
      Chen, Peter  and
      Dikkala, Nishanth  and
      Tyen, Gladys  and
      Liu, Xin  and
      Shalit, Uri  and
      Chiappa, Silvia  and
      Olszewska, Kate  and
      Tay, Yi  and
      Tran, Vinh Q.  and
      Le, Quoc V  and
      Firat, Orhan",
    editor = "Che, Wanxiang  and
      Nabende, Joyce  and
      Shutova, Ekaterina  and
      Pilehvar, Mohammad Taher",
    booktitle = "Proceedings of the 63rd Annual Meeting of the {A}ssociation for {C}omputational {L}inguistics (Volume 1: Long Papers)",
    month = jul,
    year = "2025",
    address = "Vienna, Austria",
    publisher = "Association for Computational Linguistics",
    url = "https://aclanthology.org/2025.acl-long.1285/",
    doi = "10.18653/v1/2025.acl-long.1285",
    pages = "26473--26501",
    ISBN = "979-8-89176-251-0"
}

@inproceedings{zhang2025reasoning,
    title = {Reasoning Models Know When They're Right: Probing Hidden States for Self-Verification},
    author = {Zhang, Anqi and Chen, Yulin and Pan, Jane and Zhao, Chen and Panda, Aurojit and Li, Jinyang and He, He},
    booktitle = {Second Conference on Language Modeling},
    year = {2025},
    url = {https://openreview.net/forum?id=O6I0Av7683}
}

@inproceedings{whencotfails,
    title = "When Chain-of-Thought Fails, the Solution Hides in the Hidden States",
    author = "Mehrafarin, Houman  and
      Parekh, Amit  and
      Konstas, Ioannis",
    editor = "Lee, Hung-Yi  and
      Dev, Sunipa  and
      Li, Jessy  and
      Elliott, Desmond",
    booktitle = "Findings of the {A}ssociation for {C}omputational {L}inguistics: {EMNLP} 2026",
    month = oct,
    year = "2026",
    address = "Budapest, Hungary",
    publisher = "Association for Computational Linguistics",
    url = "https://aclanthology.org/2026.findings-emnlp.2018/",
    doi = "10.18653/2026.findings-emnlp.2018",
    pages = "42939--42954",
    ISBN = "979-8-89176-462-0",
}

@inproceedings{wei2022chain,
  title     = {Chain-of-Thought Prompting Elicits Reasoning in Large Language Models},
  author    = {Wei, Jason and Wang, Xuezhi and Schuurmans, Dale and Bosma, Maarten and Ichter, Brian and Xia, Fei and Chi, Ed and Le, Quoc V and Zhou, Denny},
  booktitle = {Advances in Neural Information Processing Systems},
  volume    = {35},
  pages     = {24824--24837},
  year      = {2022},
  url       = {https://proceedings.neurips.cc/paper/2022/hash/9d5609613524ecf4f15af0f7b31abca4-Abstract-Conference.html}
}

@inproceedings{kojima2022large,
  title     = {Large Language Models are Zero-Shot Reasoners},
  author    = {Kojima, Takeshi and Gu, Shixiang (Shane) and Reid, Machel and Matsuo, Yutaka and Iwasawa, Yusuke},
  booktitle = {Advances in Neural Information Processing Systems},
  volume    = {35},
  pages     = {22199--22213},
  year      = {2022},
  url       = {https://proceedings.neurips.cc/paper/2022/hash/8bb0d291acd4acf06ef112099c16f326-Abstract-Conference.html}
}

@inproceedings{sprague2025cot,
  title     = {To {CoT} or not to {CoT}? Chain-of-thought helps mainly on math and symbolic reasoning},
  author    = {Sprague, Zayne Rea and Yin, Fangcong and Rodriguez, Juan Diego and Jiang, Dongwei and Wadhwa, Manya and Singhal, Prasann and Zhao, Xinyu and Ye, Xi and Mahowald, Kyle and Durrett, Greg},
  booktitle = {The Thirteenth International Conference on Learning Representations},
  year      = {2025},
  url       = {https://openreview.net/forum?id=w6nlcS8Kkn}
}

@inproceedings{liu2025mind,
  title     = {Mind Your Step (by Step): Chain-of-Thought can Reduce Performance on Tasks where Thinking Makes Humans Worse},
  author    = {Liu, Ryan and Geng, Jiayi and Wu, Addison J. and Sucholutsky, Ilia and Lombrozo, Tania and Griffiths, Thomas L.},
  booktitle = {Proceedings of the 42nd International Conference on Machine Learning},
  series    = {Proceedings of Machine Learning Research},
  volume    = {267},
  pages     = {38489--38517},
  publisher = {PMLR},
  year      = {2025},
  url       = {https://proceedings.mlr.press/v267/liu25t.html}
}

@inproceedings{marks2024geometry,
  title     = {The Geometry of Truth: Emergent Linear Structure in Large Language Model Representations of True/False Datasets},
  author    = {Marks, Samuel and Tegmark, Max},
  booktitle = {First Conference on Language Modeling},
  year      = {2024},
  url       = {https://openreview.net/forum?id=aajyHYjjsk}
}

@inproceedings{arditi2024refusal,
  title     = {Refusal in Language Models Is Mediated by a Single Direction},
  author    = {Arditi, Andy and Obeso, Oscar and Syed, Aaquib and Paleka, Daniel and Panickssery, Nina and Gurnee, Wes and Nanda, Neel},
  booktitle = {Advances in Neural Information Processing Systems},
  volume    = {37},
  pages     = {136037--136083},
  year      = {2024},
  url       = {https://proceedings.neurips.cc/paper_files/paper/2024/hash/f545448535dfde4f9786555403ab7c49-Abstract.html}
}

@misc{turner2024steering,
  title         = {Steering Language Models With Activation Engineering},
  author        = {Turner, Alexander Matt and Thiergart, Lisa and Leech, Gavin and Udell, David and Vazquez, Juan J. and Mini, Ulisse and MacDiarmid, Monte},
  year          = {2024},
  eprint        = {2308.10248},
  archivePrefix = {arXiv},
  primaryClass  = {cs.CL},
  url           = {https://arxiv.org/abs/2308.10248}
}

@inproceedings{li2023inference,
  title     = {Inference-Time Intervention: Eliciting Truthful Answers from a Language Model},
  author    = {Li, Kenneth and Patel, Oam and Vi{\'e}gas, Fernanda and Pfister, Hanspeter and Wattenberg, Martin},
  booktitle = {Advances in Neural Information Processing Systems},
  volume    = {36},
  pages     = {41451--41530},
  year      = {2023},
  url       = {https://proceedings.neurips.cc/paper_files/paper/2023/hash/81b8390039b7302c909cb769f8b6cd93-Abstract.html}
}

@misc{kadavath2022language,
  title         = {Language Models (Mostly) Know What They Know},
  author        = {Kadavath, Saurav and Conerly, Tom and Askell, Amanda and Henighan, Tom and Drain, Dawn and Perez, Ethan and Schiefer, Nicholas and Hatfield-Dodds, Zac and DasSarma, Nova and Tran-Johnson, Eli and Johnston, Scott and El-Showk, Sheer and Jones, Andy and Elhage, Nelson and Hume, Tristan and Chen, Anna and Bai, Yuntao and Bowman, Sam and Fort, Stanislav and Ganguli, Deep and Hernandez, Danny and Jacobson, Josh and Kernion, Jackson and Kravec, Shauna and Lovitt, Liane and Ndousse, Kamal and Olsson, Catherine and Ringer, Sam and Amodei, Dario and Brown, Tom and Clark, Jack and Joseph, Nicholas and Mann, Ben and McCandlish, Sam and Olah, Chris and Kaplan, Jared},
  year          = {2022},
  eprint        = {2207.05221},
  archivePrefix = {arXiv},
  primaryClass  = {cs.CL},
  url           = {https://arxiv.org/abs/2207.05221}
}

@inproceedings{orgad2025llms,
  title     = {{LLMs} Know More Than They Show: On the Intrinsic Representation of {LLM} Hallucinations},
  author    = {Orgad, Hadas and Toker, Michael and Gekhman, Zorik and Reichart, Roi and Szpektor, Idan and Kotek, Hadas and Belinkov, Yonatan},
  booktitle = {The Thirteenth International Conference on Learning Representations},
  year      = {2025},
  url       = {https://openreview.net/forum?id=KRnsX5Em3W}
}

@inproceedings{hojer2025improving,
  title     = {Improving Reasoning Performance in Large Language Models via Representation Engineering},
  author    = {H{\o}jer, Bertram and Jarvis, Oliver Simon and Heinrich, Stefan},
  booktitle = {The Thirteenth International Conference on Learning Representations},
  year      = {2025},
  url       = {https://openreview.net/forum?id=IssPhpUsKt}
}

@inproceedings{zhang2025uncovering,
  title     = {Uncovering Latent Chain of Thought Vectors in Large Language Models},
  author    = {Zhang, Jason and Viteri, Scott W},
  booktitle = {ICLR 2025 Workshop on Neural Network Weights as a New Data Modality},
  year      = {2025},
  url       = {https://openreview.net/forum?id=ICuIdJzBPm}
}

@inproceedings{chen2025seal,
  title     = {{SEAL}: Steerable Reasoning Calibration of Large Language Models for Free},
  author    = {Chen, Runjin and Zhang, Zhenyu and Hong, Junyuan and Kundu, Souvik and Wang, Zhangyang},
  booktitle = {Second Conference on Language Modeling},
  year      = {2025},
  url       = {https://openreview.net/forum?id=klPszYDIRT}
}

@inproceedings{azizi2025activation,
  title     = {Activation Steering for Chain-of-Thought Compression},
  author    = {Azizi, Seyedarmin and Baghaei Potraghloo, Erfan and Kundu, Souvik and Pedram, Massoud},
  editor    = {Liakata, Maria and Moreira, Viviane P. and Zhang, Jiajun and Jurgens, David},
  booktitle = {Findings of the Association for Computational Linguistics: ACL 2026},
  month     = jul,
  year      = {2026},
  address   = {San Diego, California, United States},
  publisher = {Association for Computational Linguistics},
  pages     = {36676--36687},
  doi       = {10.18653/v1/2026.findings-acl.1828},
  url       = {https://aclanthology.org/2026.findings-acl.1828/}
}

@inproceedings{meng2022locating,
  title     = {Locating and Editing Factual Associations in {GPT}},
  author    = {Meng, Kevin and Bau, David and Andonian, Alex and Belinkov, Yonatan},
  booktitle = {Advances in Neural Information Processing Systems},
  volume    = {35},
  pages     = {17359--17372},
  year      = {2022},
  url       = {https://proceedings.neurips.cc/paper_files/paper/2022/hash/6f1d43d5a82a37e89b0665b33bf3a182-Abstract-Conference.html}
}

@inproceedings{ghandeharioun2024patchscopes,
  title     = {Patchscopes: A Unifying Framework for Inspecting Hidden Representations of Language Models},
  author    = {Ghandeharioun, Asma and Caciularu, Avi and Pearce, Adam and Dixon, Lucas and Geva, Mor},
  booktitle = {Proceedings of the 41st International Conference on Machine Learning},
  series    = {Proceedings of Machine Learning Research},
  volume    = {235},
  pages     = {15466--15490},
  publisher = {PMLR},
  year      = {2024},
  url       = {https://proceedings.mlr.press/v235/ghandeharioun24a.html}
}

@inproceedings{zhang2024towards,
  title     = {Towards Best Practices of Activation Patching in Language Models: Metrics and Methods},
  author    = {Zhang, Fred and Nanda, Neel},
  booktitle = {The Twelfth International Conference on Learning Representations},
  year      = {2024},
  url       = {https://openreview.net/forum?id=Hf17y6u9BC}
}

@misc{deng2024explicit,
  title         = {From Explicit {CoT} to Implicit {CoT}: Learning to Internalize {CoT} Step by Step},
  author        = {Deng, Yuntian and Choi, Yejin and Shieber, Stuart},
  year          = {2024},
  eprint        = {2405.14838},
  archivePrefix = {arXiv},
  primaryClass  = {cs.CL},
  url           = {https://arxiv.org/abs/2405.14838}
}

@inproceedings{hao2025training,
  title     = {Training Large Language Models to Reason in a Continuous Latent Space},
  author    = {Hao, Shibo and Sukhbaatar, Sainbayar and Su, DiJia and Li, Xian and Hu, Zhiting and Weston, Jason E and Tian, Yuandong},
  booktitle = {Second Conference on Language Modeling},
  year      = {2025},
  url       = {https://openreview.net/forum?id=Itxz7S4Ip3}
}

@inproceedings{goyal2024think,
  title     = {Think before you speak: Training Language Models With Pause Tokens},
  author    = {Goyal, Sachin and Ji, Ziwei and Rawat, Ankit Singh and Menon, Aditya Krishna and Kumar, Sanjiv and Nagarajan, Vaishnavh},
  booktitle = {The Twelfth International Conference on Learning Representations},
  year      = {2024},
  url       = {https://openreview.net/forum?id=ph04CRkPdC}
}

@inproceedings{you2026spherical,
  title     = {Spherical Steering: Geometry-Aware Activation Rotation for Language Models},
  author    = {You, Zejia and Deng, Chunyuan and Chen, Hanjie},
  booktitle = {Proceedings of the 43rd International Conference on Machine Learning},
  series    = {Proceedings of Machine Learning Research},
  volume    = {306},
  pages     = {149801--149823},
  publisher = {PMLR},
  year      = {2026},
  url       = {https://proceedings.mlr.press/v306/you26a.html}
}

@misc{llama3,
  title         = {The {Llama} 3 Herd of Models},
  author        = {Grattafiori, Aaron and Dubey, Abhimanyu and Jauhri, Abhinav and Pandey, Abhinav and Kadian, Abhishek and Al-Dahle, Ahmad and Letman, Aiesha and Mathur, Akhil and Schelten, Alan and Vaughan, Alex and Yang, Amy and Fan, Angela and Goyal, Anirudh and Hartshorn, Anthony and Yang, Aobo and Mitra, Archi and Sravankumar, Archie and Korenev, Artem and Hinsvark, Arthur and others},
  year          = {2024},
  eprint        = {2407.21783},
  archivePrefix = {arXiv},
  primaryClass  = {cs.AI},
  url           = {https://arxiv.org/abs/2407.21783}
}

@misc{qwen,
  title         = {{Qwen2.5} Technical Report},
  author        = {{Qwen} and Yang, An and Yang, Baosong and Zhang, Beichen and Hui, Binyuan and Zheng, Bo and Yu, Bowen and Li, Chengyuan and Liu, Dayiheng and Huang, Fei and Wei, Haoran and Lin, Huan and Yang, Jian and Tu, Jianhong and Zhang, Jianwei and Yang, Jianxin and Yang, Jiaxi and Zhou, Jingren and Lin, Junyang and Dang, Kai and Lu, Keming and Bao, Keqin and Yang, Kexin and Yu, Le and Li, Mei and Xue, Mingfeng and Zhang, Pei and Zhu, Qin and Men, Rui and Lin, Runji and Li, Tianhao and Tang, Tianyi and Xia, Tingyu and Ren, Xingzhang and Ren, Xuancheng and Fan, Yang and Su, Yang and Zhang, Yichang and Wan, Yu and Liu, Yuqiong and Cui, Zeyu and Zhang, Zhenru and Qiu, Zihan},
  year          = {2025},
  eprint        = {2412.15115},
  archivePrefix = {arXiv},
  primaryClass  = {cs.CL},
  url           = {https://arxiv.org/abs/2412.15115}
}

@misc{gsm8k,
  title         = {Training Verifiers to Solve Math Word Problems},
  author        = {Cobbe, Karl and Kosaraju, Vineet and Bavarian, Mohammad and Chen, Mark and Jun, Heewoo and Kaiser, Lukasz and Plappert, Matthias and Tworek, Jerry and Hilton, Jacob and Nakano, Reiichiro and Hesse, Christopher and Schulman, John},
  year          = {2021},
  eprint        = {2110.14168},
  archivePrefix = {arXiv},
  primaryClass  = {cs.LG},
  url           = {https://arxiv.org/abs/2110.14168}
}

@inproceedings{bbh,
    title = "Challenging {BIG}-Bench Tasks and Whether Chain-of-Thought Can Solve Them",
    author = {Suzgun, Mirac  and
      Scales, Nathan  and
      Sch{\"a}rli, Nathanael  and
      Gehrmann, Sebastian  and
      Tay, Yi  and
      Chung, Hyung Won  and
      Chowdhery, Aakanksha  and
      Le, Quoc  and
      Chi, Ed  and
      Zhou, Denny  and
      Wei, Jason},
    editor = "Rogers, Anna  and
      Boyd-Graber, Jordan  and
      Okazaki, Naoaki",
    booktitle = "Findings of the Association for Computational Linguistics: ACL 2023",
    month = jul,
    year = "2023",
    address = "Toronto, Canada",
    publisher = "Association for Computational Linguistics",
    url = "https://aclanthology.org/2023.findings-acl.824/",
    doi = "10.18653/v1/2023.findings-acl.824",
    pages = "13003--13051",
}

@inproceedings{paszke2019pytorch,
  title     = {{PyTorch}: An Imperative Style, High-Performance Deep Learning Library},
  author    = {Paszke, Adam and Gross, Sam and Massa, Francisco and Lerer, Adam and Bradbury, James and Chanan, Gregory and Killeen, Trevor and Lin, Zeming and Gimelshein, Natalia and Antiga, Luca and Desmaison, Alban and K{\"o}pf, Andreas and Yang, Edward and DeVito, Zachary and Raison, Martin and Tejani, Alykhan and Chilamkurthy, Sasank and Steiner, Benoit and Fang, Lu and Bai, Junjie and Chintala, Soumith},
  booktitle = {Advances in Neural Information Processing Systems},
  volume    = {32},
  year      = {2019},
  url       = {https://proceedings.neurips.cc/paper/2019/hash/bdbca288fee7f92f2bfa9f7012727740-Abstract.html}
}

@inproceedings{wolf-etal-2020-transformers,
    title = "Transformers: State-of-the-Art Natural Language Processing",
    author = "Wolf, Thomas  and Debut, Lysandre  and Sanh, Victor  and Chaumond, Julien  and Delangue, Clement  and Moi, Anthony  and Cistac, Pierric  and Rault, Tim  and Louf, Remi  and Funtowicz, Morgan  and Davison, Joe  and Shleifer, Sam  and von Platen, Patrick  and Ma, Clara  and Jernite, Yacine  and Plu, Julien  and Xu, Canwen  and Le Scao, Teven  and Gugger, Sylvain  and Drame, Mariama  and Lhoest, Quentin  and Rush, Alexander",
    editor = "Liu, Qun  and Schlangen, David",
    booktitle = "Proceedings of the 2020 Conference on Empirical Methods in Natural Language Processing: System Demonstrations",
    month = oct,
    year = "2020",
    address = "Online",
    publisher = "Association for Computational Linguistics",
    url = "https://aclanthology.org/2020.emnlp-demos.6/",
    doi = "10.18653/v1/2020.emnlp-demos.6",
    pages = "38--45"
}

@article{virtanen2020scipy,
    title = {{SciPy} 1.0: fundamental algorithms for scientific computing in {P}ython},
    author = {Virtanen, Pauli and Gommers, Ralf and Oliphant, Travis E. and Haberland, Matt and Reddy, Tyler and Cournapeau, David and Burovski, Evgeni and Peterson, Pearu and Weckesser, Warren and Bright, Jonathan and van der Walt, St{\'e}fan J. and Brett, Matthew and others},
    journal = {Nature Methods},
    volume = {17},
    number = {3},
    pages = {261--272},
    year = {2020},
    doi = {10.1038/s41592-019-0686-2}
}
